\documentclass[11pt]{article}
\usepackage{iftex}
\ifPDFTeX\usepackage[T1]{fontenc}\usepackage[utf8]{inputenc}\fi
\usepackage[margin=1in]{geometry}
\usepackage{graphicx}
\usepackage{array}
\usepackage{amsmath,amssymb}
\usepackage{hyperref}
\usepackage{xurl}
\hypersetup{colorlinks=true,linkcolor=blue,urlcolor=blue}
\usepackage{parskip}
\graphicspath{{paper_v2/figures/}{figures/}}
\begin{document}
\title{AgentTrust: A Self-Improving Trust Layer for AI-Agent Actions}\author{Chenglin Yang\\\textit{Independent Researcher}\\\texttt{yangchenglin802@gmail.com}}\date{June 2026}\maketitle
*AgentTrust v2 --- a trust layer for agent actions that \textbf{learns from its own decisions}: it distils rules on lexical threats (cheaper over time) and accrues guarded memory on semantic threats (smarter over time).*

\subsection*{Abstract}
AI agents increasingly take consequential actions---shell commands, cloud operations, and arbitrary tool-calls---so a trust layer must decide, per action, whether to allow, warn, block, or escalate. We argue that the right way to reason about such a layer is by \textit{threat type}. \textbf{Lexical (fixed-signature) threats}, where danger lives in a stable token, are decidable by deterministic rules; \textbf{semantic (intent-dependent) threats}, where a benign and a malicious action share the same surface, are out of reach for rules \textit{by construction}. We make this concrete with a negative proof: a determined, hand-authored cloud rule pack lifts held-out accuracy only 48$\rightarrow$56\% overall and moves the semantic categories by 0pp (data\_db 29$\rightarrow$29, observability 59$\rightarrow$59, supply\_chain 50$\rightarrow$50), while a strong LLM judge carries exactly those categories. We give the judge a self-learning capability---on a corpus that is mainly semantic attacks it nearly doubles rule accuracy (48\% $\rightarrow$ 83.6--85.2\%) with near-zero false-blocks, and this holds across two model providers. We turn this into a \textbf{self-improving dual-store system}: the judge distills a growing deterministic rule floor on lexical threats (cheaper over time) and feeds a \textit{guarded} RAG memory on semantic threats (a verdict-cache fails---surface-twins collapse to \textasciitilde{}58\%---so a corroboration guard lifts semantic accuracy +13pp, 70$\rightarrow$84). The result is what sets AgentTrust v2 apart from its \textit{static} v1 predecessor: a trust layer that self-evolves from its own stream of decisions --- getting cheaper on the lexical class (it distils its own rules) and smarter on the semantic class (it accrues guarded precedent), while never hard-blocking a benign action. An end-to-end online replay shows the judge-call rate falling (50\%$\rightarrow$44\%) and judge-domain accuracy rising (71\%$\rightarrow$80\%) over a traffic stream, with 0 benign hard-blocks across 45,000 actions.

\section*{1. Introduction}
AI agents no longer merely produce text; they \textit{act}. Modern agents complete tasks by emitting tool-calls --- shell commands, HTTP requests, database queries, cloud-control-plane operations, package installs --- that change the state of the world. Each such action carries a real blast radius: a single tool-call can delete a filesystem, exfiltrate secrets to an attacker-controlled endpoint, grant a persistent IAM credential, or pull a backdoored dependency. As agents are wired into shells, CI/CD, and cloud accounts, the unit that must be trusted shifts from the \textit{agent} to the \textit{individual action}. This motivates a \textbf{per-action trust layer}: a guardrail that intercepts each tool-call and returns a verdict --- allow, warn, block, or escalate to human review --- before the action executes.

The dominant approach to such guardrails is deterministic policy: regexes, AST matchers, and allow/deny lists that flag known-bad patterns. These rules are fast, cheap, auditable, and high-precision on the threats they were written for. But they share a structural limitation that, we argue, no amount of rule-engineering can overcome --- and it is best seen by decomposing agent threats \textit{by type} rather than by domain.

\subsection*{A threat-type decomposition: lexical vs. semantic}
We distinguish two kinds of dangerous action. A \textbf{lexical} (fixed-signature) threat is one whose danger lives in a stable, surface-level token: \texttt{rm -rf /\allowbreak{}}, a reverse-shell one-liner, a hardcoded private key, \texttt{iam create-access-key}, \texttt{--acl public-read}, a cluster-admin binding, a write to \texttt{ld.so.preload}, a request to the cloud metadata IP. The malicious action \textit{looks} malicious; a regex can decide it. A \textbf{semantic} (intent-dependent) threat is one where the benign and the malicious action \textbf{share the same surface}, and the verdict turns on destination, content, or scope rather than on any token: a telemetry \texttt{curl} versus an exfiltration \texttt{curl} (including abuse of an \textit{allow-listed} endpoint); a \texttt{kubectl get secret} issued for debugging versus for theft; a read-only \texttt{SELECT} versus a \texttt{COPY ... TO} that dumps a table to an external sink; a legitimate dependency versus one whose \texttt{postinstall} hook executes code. For semantic threats, the surface is identical across the two verdicts, so any regex precise enough to block the attack must also block its benign twin --- i.e., rules cannot separate them \textit{by construction}.

\begin{figure}[ht]\centering
\includegraphics[width=\linewidth]{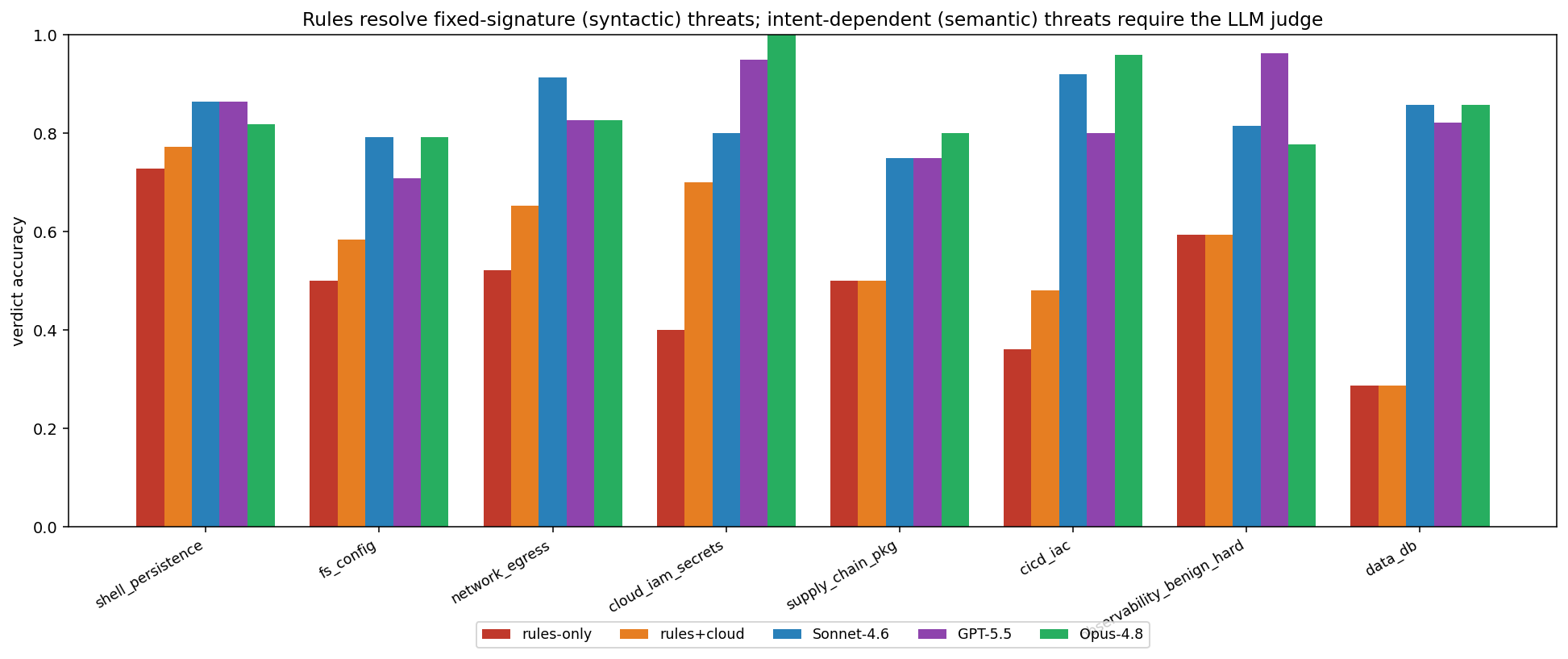}
\caption{Held-out external corpus: deterministic rules close on lexical (fixed-signature) threat categories and plateau on the semantic (intent-dependent) ones, while strong LLM judges across two providers carry exactly the semantic categories.}\label{fig:fig1}
\end{figure}

This is not a hypothesis about under-investment in rule-writing; it is a claim about a structural ceiling. We establish it with a determined \textbf{steelman}. Starting from an honest deterministic rule pack, we hand-authored an additional cloud/CI/Kubernetes rule pack \textit{on development data} and tested it on a held-out, realistic corpus. The added rules lift overall held-out accuracy from 48\% to 56\% --- exactly the kind of gain one expects when more lexical signatures are added --- yet they move the purely semantic categories by 0 percentage points: data-and-database threats stay at 29$\rightarrow$29, observability/benign-hard cases at 59$\rightarrow$59, and supply-chain threats at 50$\rightarrow$50 (Table 3). A separate, maximally token-tuned "cheat" rule set, optimized directly against the data, also lands at 47.6\% overall: in-distribution tuning does not transfer. Rules close on the lexical categories and plateau, untouched, on the semantic ones. Figure~\ref{fig:fig1} makes this concrete: rules climb on fixed-signature categories and flatline on intent-dependent ones, while a strong LLM judge carries precisely those semantic categories (78--100\%).

\subsection*{The judge handles what rules cannot --- but its trade-offs are uncharacterized}
If rules cannot cross the semantic boundary, an obvious candidate is an LLM-as-judge that reads the action and reasons about intent. LLM-based guardrails already exist. The open question --- and the one this paper answers --- is not \textit{whether} to use a judge, but \textit{how it behaves as a system component}: its precision/recall trade-off against rules, its cost, its robustness across model families and providers, and whether it can be made to \textbf{learn} rather than re-pay full inference cost forever. These properties are largely uncharacterized in prior work, and they matter because a naive judge can be worse than rules: judging every action invites false-positives and unbounded cost.

We characterize the judge across three corpora --- an internal tuning set, an independent out-of-sample set, and a realistic held-out corpus independently relabeled by a second annotator ($\kappa$=0.82) --- and across four model families and two providers. The headline is that the judge handles the semantic threats rules cannot, and we make it self-learning. On a mainly-semantic-attack corpus, deterministic rules collapse to 48.1\% accuracy, while four judges spanning two providers cluster at \textbf{83.6--85.2\%} --- nearly doubling rule accuracy --- with near-zero false-blocks (Table 1). In-distribution, where rules are already strong, a \textit{confidence-gated} judge that escalates only when it is confident is net-positive on every axis: gating the judge atop the rule floor reaches 98.3/2.1/0.9 (accuracy/FPR/FNR) on the independent 630-action set, versus 95.4/2.1/6.1 for rules alone, while holding the false-positive rate fixed (Table 2). Crucially, recall \textit{saturates} even at the smallest model we test --- every judge catches the semantic attacks rules miss --- so additional model capability buys \textbf{precision}, not recall. And across every judge and every dataset, the system records \textbf{zero benign hard-blocks}: the worst-case precision cost of escalation is a soft warning, never a hard block on a legitimate action. This safety invariant is what makes the layer deployable.

\subsection*{From a better judge to a system that learns}
\begin{figure}[ht]\centering
\includegraphics[width=\linewidth]{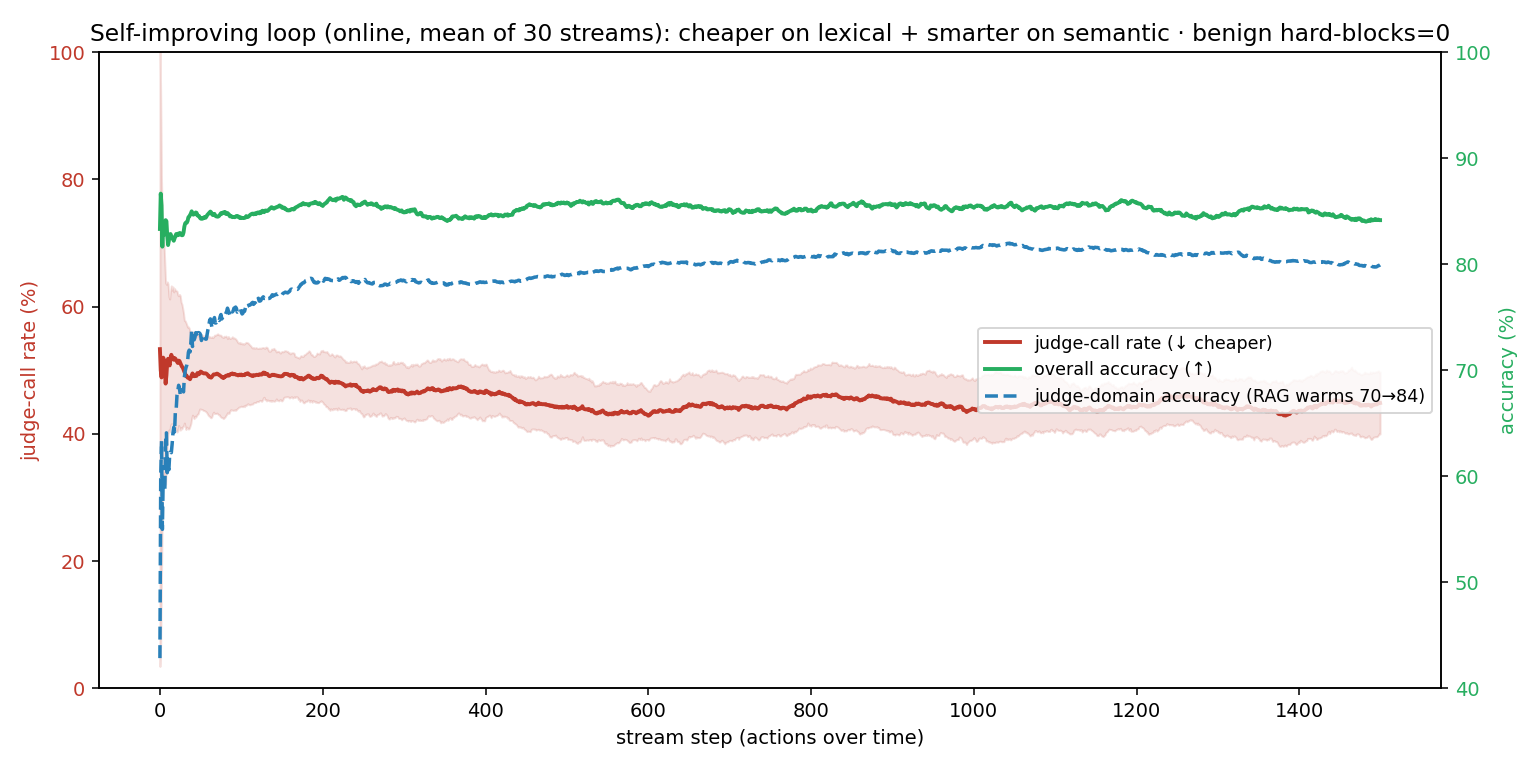}
\caption{Online self-improving loop (mean of 30 replayed streams): the judge-call rate falls (cheaper, via lexical rule distillation) while judge-domain accuracy rises (smarter, as the guarded RAG memory warms); 0 benign hard-blocks throughout.}\label{fig:fig2}
\end{figure}

\begin{figure}[ht]\centering
\includegraphics[width=\linewidth]{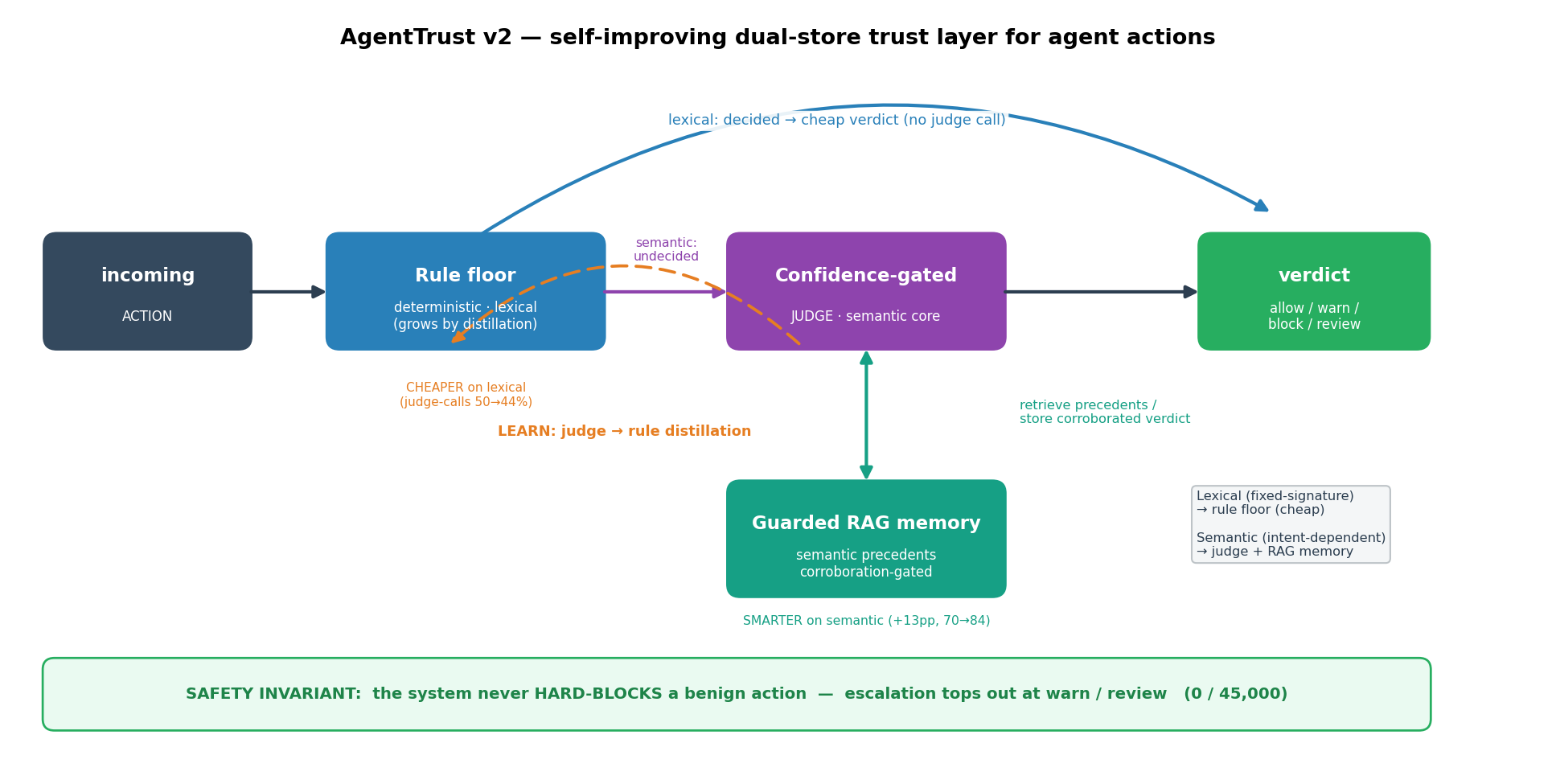}
\caption{AgentTrust v2 architecture: a deterministic rule floor (lexical) and a confidence-gated judge backed by a guarded RAG memory (semantic), with the judge-as-teacher learning loop and the never-hard-block-a-benign-action safety invariant.}\label{fig:fig4}
\end{figure}

A static judge that generalizes is useful but expensive and forgetful. \textbf{The defining capability of AgentTrust v2 --- and its central departure from the \textit{static} v1 interceptor --- is that it self-evolves from its own stream of judgments.} The novelty of this work is to turn the threat-type decomposition into a self-improving dual-store system in which the judge acts as a \textit{teacher} for two different students, each matched to one threat type --- getting \textbf{cheaper on the lexical} and smarter on the semantic over time, while preserving the never-hard-block-a-benign-action invariant. Figure~\ref{fig:fig2} sketches the loop, and Figure~\ref{fig:fig4} gives the architecture: an action passes through a deterministic \textbf{rule floor}; lexically decidable cases get a cheap verdict, and undecided cases are sent to the \textbf{confidence-gated judge} augmented by a \textbf{guarded retrieval memory}; the system then \textit{learns} by distilling new lexical rules and storing semantic precedents.

\begin{figure}[ht]\centering
\includegraphics[width=\linewidth]{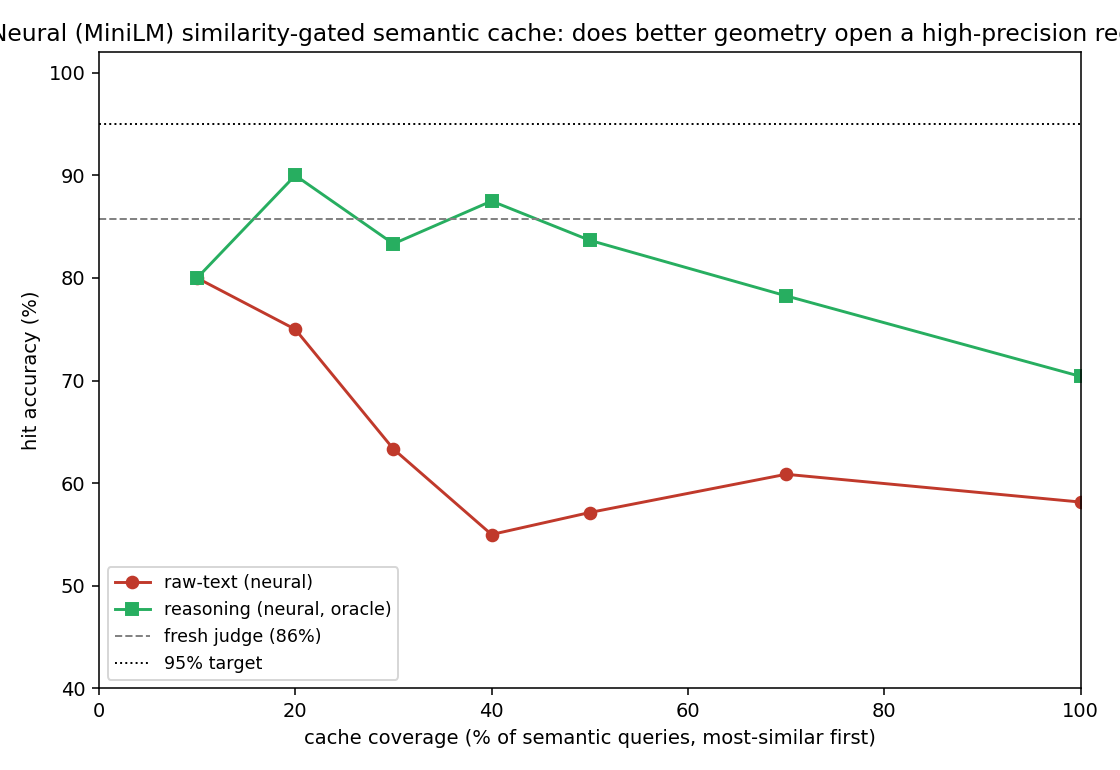}
\caption{A similarity-gated semantic verdict-cache has no high-precision regime at any coverage (neither raw-text nor neural retrieval) because surface-twins collapse -- which is why the semantic store must be retrieval-augmented generation, not a cache.}\label{fig:fig3}
\end{figure}

The two students are deliberately asymmetric, because the two threat types are. For \textbf{lexical} threats, the judge distills new deterministic rules, so a growing rule floor handles more traffic without a judge call --- the judge-call rate falls over a traffic stream. For \textbf{semantic} threats, distillation is impossible (there is no stable signature to distill), so the judge instead populates a retrieval-augmented memory of precedents. Here we report a sharp \textbf{negative} result that dictates the design: a semantic verdict-\textit{cache} keyed on the action cannot work, because retrieval collapses surface-twins --- raw and neural retrieval both sit near 58\% accuracy with \textbf{no high-precision regime at any coverage} (Figure~\ref{fig:fig3}). The memory must therefore be RAG, not a cache: precedents augment the judge's reasoning rather than replace its verdict. Naively storing the judge's own verdicts as precedents barely helps (+2pp) and can poison a category; gating admission on \textbf{corroboration} between two judges raises semantic accuracy by +13pp (70$\rightarrow$84) with zero dangerous leaks, lifting memory correctness from 70\% to \textbf{97\%} (Table 4). End to end, an online replay over many traffic streams shows the loop working as intended: the judge-call rate falls from 50\% to \textbf{44\%} (cheaper), judge-domain accuracy on the non-rule-able semantic cases rises from 71\% to \textbf{80\%} (smarter), and the system commits \textbf{0 benign hard-blocks across 45,000 actions} (Table 5).

\subsection*{Contributions}
This paper makes three contributions; the third --- \textbf{a trust layer that learns and evolves from its own judgments} --- is the centerpiece, and the first two are the decomposition and characterization that make that self-evolution well-typed.

\begin{enumerate}
\item \textbf{A threat-type decomposition with a negative proof.} We separate agent-action threats into \textit{lexical} (fixed-signature, rule-decidable) and \textit{semantic} (intent-dependent, where benign and malicious share a surface) classes, and we show --- via a determined steelman --- that rules are surface-pattern-bound: a hand-authored cloud rule pack lifts held-out accuracy 48$\rightarrow$56\% overall but moves every semantic category 0 percentage points (data\_db 29$\rightarrow$29, observability 59$\rightarrow$59, supply\_chain 50$\rightarrow$50), and a max-tuned rule set still reaches only 47.6\%. The semantic boundary is structural, not an artifact of insufficient rules (Figure~\ref{fig:fig1}).
\end{enumerate}
\begin{enumerate}
\item \textbf{A cross-distribution $\times$ cross-provider characterization of the judge.} We show that a strong LLM judge is the component that generalizes where rules do not: on a realistic held-out corpus, rules score 48.1\% while four judges across two providers score 83.6--85.2\% with near-zero false-blocks, and in-distribution a \textit{confidence-gated} judge is net-positive on every axis (98.3/2.1/0.9 vs. 95.4/2.1/6.1). We further establish two design-relevant invariants: judge recall \textit{saturates} (capability buys precision, not recall) and the system records \textbf{zero benign hard-blocks} across every judge and dataset.
\end{enumerate}
\begin{enumerate}
\item \textbf{A self-improving dual-store system.} We turn the decomposition into a system in which the judge teaches a growing deterministic rule floor on lexical threats (cheaper over time) and a \textit{guarded} RAG memory on semantic threats (smarter over time). We report the negative result that a semantic verdict-cache cannot work (retrieval collapses surface-twins; no high-precision regime) --- which is \textit{why} the semantic store must be retrieval-augmented --- and the positive result that a corroboration-guarded memory lifts semantic accuracy +13pp (70$\rightarrow$84) and memory correctness to 97\%. An end-to-end online simulation demonstrates the loop: judge-cost falls (50$\rightarrow$44\% call rate), judge-domain accuracy rises (71$\rightarrow$80\%), and \textbf{0 benign actions are hard-blocked over 45,000 actions} (Figure~\ref{fig:fig2}).
\end{enumerate}
\textbf{The v1 $\rightarrow$ v2 delta, in three points.} v1 [Yang 2026] was a \textit{static} interceptor: its deterministic rules carried the headline numbers and its judge was opt-in --- indeed a naive always-on judge \textit{lowered} accuracy (to 88.0\%, false positives 2.3$\rightarrow$ 9.0\%). v2 changes exactly three things. (i) It shows the rules-vs-judge split is \textit{structural}, not a coverage gap --- a determined rule steelman moves the semantic categories by 0pp, so a judge is necessary, not optional. (ii) It turns the judge into a \textit{net-positive, confidence-gated} core that holds across distributions and two providers --- the asymmetric design v1 explicitly left to future work. (iii) It makes the layer \textit{self-evolving}: it distils cheap rules on lexical threats and accrues a guarded retrieval memory on semantic threats from its own decisions.

The remainder of the paper makes these claims precise. We position against rule/policy guardrails and LLM-as-judge guardrails (\S{}2), formalize the action representation and threat taxonomy (\S{}3), describe the dual-store system and its safety invariant (\S{}4), and evaluate all three pillars across three datasets and five model configurations (\S{}5). We are deliberate about scope: our evaluation runs on curated corpora through an agent/replay harness rather than live production traffic, and the online loop is a measured replay simulation rather than a deployed service (\S{}6). We state plainly the claims we do \textit{not} make --- in particular, we do not claim the judge alone beats rules in-distribution on exact-match, nor that adding more rules is a contribution (\S{}6).

\section*{2. Related Work}
AgentTrust v2 sits at the intersection of three lines of work: (a) rule- and policy-based \textit{static} guardrails, (b) \textit{model-based} guardrails that use a language model to judge inputs, outputs, or actions, and (c) the machine-learning machinery for making such a system \textit{learn over time} --- knowledge distillation, retrieval-augmented generation (RAG), case-based reasoning (CBR), and online learning. We review each in turn and, in each case, state precisely where our contribution lies. We stress at the outset what our contribution is \textbf{not}: it is not "use an LLM as a judge." LLM-as-judge is by now a standard tool [Zheng et al. 2023], and a prior version of this very system already contained an (opt-in) judge [Yang 2026]. Our claims are three structural ones that the literature below does not make: \textbf{(1)} a decomposition of agent- action threats into a \textit{lexical / fixed-signature} class and a \textit{semantic / intent-dependent} class, together with a \textbf{negative result} that deterministic rules cannot cross the boundary regardless of how many rules are added; \textbf{(2)} a \textbf{cross-distribution $\times$ cross-provider} characterization showing the judge --- not the rules --- is the component that \textit{generalizes}; and \textbf{(3)} a \textbf{self-improving dual-store} in which the judge acts as a \textit{teacher} that distills a fast deterministic rule "student" on the lexical class and curates a \textit{guarded} retrieval memory on the semantic class, under a never-hard-block-a-benign-action safety invariant.

\subsection*{2.1 Rule- and policy-based static guardrails}
The first and still most widely deployed line of defense for consequential software actions is the deterministic policy engine. In cloud and Kubernetes infrastructure, \textbf{Open Policy Agent (OPA)} and its Rego policy language are the de facto standard: policies are written as code, versioned, and consulted at admission time to allow or deny an operation [Open Policy Agent]. For LLM applications specifically, \textbf{NeMo Guardrails} [Rebedea et al. 2023] provides \textit{programmable rails} --- user-authored, model-independent, interpretable flows in the Colang language --- that constrain dialogue and tool use. Lightweight pattern-and-signature scanners such as \textbf{Rebuff} and \textbf{Vigil} detect known-bad inputs via heuristics, regexes, and embedding-similarity to a database of past attacks [Rebuff; Vigil]. The v1 of AgentTrust [Yang 2026] is itself in this family for its core path: a deterministic shell-deobfuscation normalizer plus hand-authored rules that flag fixed dangerous signatures (\texttt{rm -rf /\allowbreak{}}, reverse shells, hardcoded keys), with an LLM judge bolted on only for ambiguous inputs.

All of these share a structural property: a rule fires on a \textbf{surface pattern}. This is exactly the right tool for what we call \textit{lexical / fixed-signature} threats, where the danger is a stable token. It is the wrong tool --- \textit{by construction}, not merely by under-engineering --- for \textit{semantic / intent- dependent} threats, where a benign and a malicious action share an identical surface and differ only in destination, scope, or content (telemetry \texttt{curl} vs. exfiltration \texttt{curl} to an attacker host; \texttt{kubectl get secret} for debugging vs. for theft; \texttt{SELECT} vs. a \texttt{COPY}-to-exfil). A pattern that matches the malicious instance necessarily matches its benign twin, so any rule precise enough to catch the attack must false-positive on legitimate traffic. \textbf{This is our Pillar 2, and we believe it is novel as a \textit{demonstrated} result.} Prior work notes informally that static filters "miss obfuscation and multi-step context" [Yang 2026] or are brittle, but does not isolate \textit{which} threats are rule-decidable and \textit{which} are not, nor prove the gap survives a determined fix. We supply a steelman: a hand-authored cloud/Kubernetes/CI rule pack, authored on development sets and tested held-out, lifts overall held-out accuracy only 48$\rightarrow$56\% and moves every semantic category 0pp (data/DB 29$\rightarrow$29, observability 59$\rightarrow$59, supply-chain 50$\rightarrow$50), while a max-tuned "cheat" rule set also plateaus. The boundary is structural; you cannot regex your way across it.

\subsection*{2.2 Model-based guardrails and LLM-as-judge}
A second line replaces (or augments) hand-written patterns with a learned model. \textbf{Llama Guard} [Inan et al. 2023] is an instruction-tuned Llama-2-7B classifier that labels prompts and responses against a safety taxonomy; \textbf{Llama Guard 3} extends this with the Llama-3 release [Grattafiori et al. 2024]. \textbf{Meta Prompt Guard} is a smaller mDeBERTa-based classifier specialized to prompt-injection and jailbreak detection [Meta Prompt Guard], as is ProtectAI's widely used DeBERTa-v3 prompt-injection detector [ProtectAI]. More broadly, \textbf{constitutional AI} [Bai et al. 2022] uses a model to critique and revise outputs against a set of written principles, and the \textbf{LLM-as-a-judge} paradigm [Zheng et al. 2023] established strong models as evaluators of open-ended text. Surveys of this rapidly growing "guardrail" space catalog input/output rails, classifiers, and detectors [Dong et al. 2024].

A closely related and very active sub-area targets \textbf{agentic} safety specifically. Greshake et al. [2023] introduced \textit{indirect} prompt injection against tool-using, LLM-integrated applications; \textbf{AgentDojo} [Debenedetti et al. 2024] and \textbf{ToolEmu} [Ruan et al. 2024] are benchmarks/sandboxes that \textit{measure} agent vulnerability and risky behavior; \textbf{CaMeL} [Debenedetti et al. 2025], \textbf{Progent} [Shi et al. 2025], and \textbf{AgentBound} [Bühler et al. 2026] \textit{defend} by enforcing capability/control-flow, programmable-privilege, or execution-boundary constraints around tool calls; and the \textbf{OWASP Top 10 for LLM Applications} [OWASP 2025] codifies prompt injection, improper output handling, and "excessive agency" as headline risks. These works establish the \textit{threat} (agents take consequential actions over untrusted data) and supply \textit{runtime} defenses, motivating a per-action trust layer like ours.

\textbf{Where we differ.} Our contribution is emphatically \textit{not} the use of a model as a judge. The components above are, with few exceptions, evaluated in \textit{one} setting and treated as static artifacts, and three gaps remain that we close:

\begin{enumerate}
\item \textbf{Threat-type decidability.} None of these works partition the threat space into a rule-decidable class and a rule-undecidable class and \textit{prove} (via a determined steelman) that rules cannot cover the latter --- the core conceptual move that \textit{justifies} needing a judge in the first place rather than just asserting it (\S{}2.1, Pillar 2).
\item \textbf{Generalization across distribution and provider.} Classifier guardrails such as Llama Guard, Prompt Guard, and DeBERTa detectors are themselves trained models and are known to degrade off-distribution and under simple evasions [Meta Prompt Guard; Dong et al. 2024]. We instead characterize, on a realistic held-out (out-of-distribution) corpus, that hand-authored \textit{rules} collapse (48\% accuracy) while a \textit{strong general} judge nearly doubles them ($\approx$84--85\%) with near-zero false-blocks, and --- critically --- that this holds across two model providers and a capability ladder (recall saturates even at a small model; capability buys \textit{precision}). This is Pillar 1, and it is a \textit{measurement/positioning} result, not a new model.
\item \textbf{The cost/precision trade-off and the safety invariant.} We do not claim the judge \textit{alone} beats rules in-distribution on exact-match (it does not). Our in-distribution result is a \textbf{confidence-gated} hybrid that is net-positive, with a \textit{provider-dependent} gate threshold (load-bearing), and a hard invariant that the system never \textit{hard-blocks} a benign action (escalation tops out at warn/review). Existing guardrail papers report aggregate accuracy/ASR; they do not characterize a gated cost-vs-precision frontier under a no-false-hard-block guarantee.
\end{enumerate}
In short, prior model-guardrail work answers "\textit{can} a model flag unsafe content?" We answer "\textit{which} threats need a model at all, \textit{does} the model generalize where rules don't (across distributions and providers), and \textit{how} do we deploy it cheaply and safely."

\subsection*{2.3 Distillation, retrieval/case-based reasoning, and online learning}
Our third contribution --- the self-improving dual-store --- recombines three classical ML ideas. We frame the judge as a \textbf{teacher} in the sense of \textit{knowledge distillation} [Hinton et al. 2015], which transfers a large model's behavior into a cheaper student. Unlike standard distillation into a \textit{neural} student, our lexical student is a growing set of \textit{deterministic rules}: the judge labels lexically-novel attacks, and confirmed fixed-signature verdicts are promoted to exact/regex rules, so the system gets \textbf{cheaper over time} on the lexical class. This is closest in spirit to \textbf{symbolic knowledge distillation} [West et al. 2021], which distills a large LM into an explicit symbolic artifact (there, a commonsense knowledge graph) rather than weights --- but applied to a \textit{safety policy} and gated for precision so the distilled rules never inherit the teacher's false positives.

For the \textit{semantic} class --- which \S{}2.1 argues is rule-undecidable \textit{permanently} --- we instead keep a \textit{retrieval} memory of precedents, in the lineage of \textbf{retrieval-augmented generation} [Lewis et al. 2020] and, conceptually, the much older \textbf{case-based reasoning} paradigm [Aamodt and Plaza 1994], which solves a new problem by retrieving and adapting solutions to similar past cases. The judge is augmented at decision time with the most similar adjudicated precedents, so the system gets \textbf{smarter over time} on semantics. Two findings make this more than a textbook application of RAG/CBR. First, a \textit{negative} one: a naive \textit{verdict-cache} (retrieve the nearest case and reuse its label) \textbf{cannot work} here, because the very surface-twin property that defeats rules also collapses semantic twins under both lexical and neural retrieval (no high-precision regime; $\approx$58\% accuracy) --- which is \textit{why} the semantic store must be a judge-augmenting RAG, not a cache. Second, a \textit{guardedness} one: admitting raw judge verdicts into memory poisons it (a benign-looking exfiltration precedent can flip a later twin), so we add a \textbf{corroboration gate} --- admit a precedent only when two judges agree --- which lifts semantic accuracy +13pp (70$\rightarrow$84) and keeps memory $\approx$97\% correct with zero dangerous leaks. The closest analogues (self-hardening detectors that learn from caught attacks, e.g., Rebuff's growing attack vector store [Rebuff]) accumulate \textit{attack} signatures into a similarity index; they do not (i) split the store by threat-decidability, (ii) report the cache-fails-for-semantics negative result, or (iii) impose a corroboration guard to keep a \textit{learned} memory poison-resistant under a safety invariant.

Finally, the end-to-end behavior --- a policy that improves from its own stream of decisions --- places the system in the tradition of \textbf{online learning} [Shalev- Shwartz 2012], where a predictor updates from a sequence of feedback. Our online contribution is empirical and systems-flavored rather than a new regret bound: a replay simulation over a traffic stream shows the judge-call rate \textbf{falling} (50$\rightarrow$44\%, the distillation effect) while judge-domain accuracy \textbf{rises} (71$\rightarrow$80\%, the memory effect), with 0 benign hard-blocks across 45,000 actions. To our knowledge no prior agent-guardrail system demonstrates this dual cheaper-on- lexical / smarter-on-semantic dynamic, nor ties it to the threat-type decomposition that makes each half well-typed (rules for the distillable lexical threats, guarded memory for the non-distillable semantic ones).

\subsection*{2.4 Relationship to AgentTrust v1}
This paper evolves AgentTrust v1 [Yang 2026], which introduced the per-action allow/warn/block/review trust layer, the shell-deobfuscation normalizer, SafeFix, RiskChain multi-step detection, and a cache-aware LLM-as-judge exposed over MCP. In v1 the deterministic rules carried the headline numbers and the judge was effectively opt-in. v2 makes the judge the evaluated core and contributes what v1 lacked: the threat-type decidability proof (\S{}2.1), the cross-distribution $\times$ cross-provider generalization characterization (\S{}2.2), and the self-improving guarded dual-store with its negative cache result and corroboration guard (\S{}2.3).

Two points sharpen this lineage --- and the leap from v1. First, v1's judge was \textit{opt-in}, and v1's own \textit{naive} hybrid (consult the judge on every non-block action and take the more severe verdict) actually \textbf{hurt}: it cut misses but pushed the false-positive rate from 2.3\% to 9.0\% and dropped aggregate accuracy to 88.0\% [Yang 2026, Table 9] --- the ``judge hurts'' result. v1 anticipated the fix, explicitly previewing an \textit{asymmetric} hybrid (let the judge escalate only when the rule verdict is itself uncertain) as future work [Yang 2026, \S{}6.5]; v2's confidence-gated, escalate-only judge \textbf{is} that design, now delivered, characterized, and shown net-positive across providers (\S{}5.2). Second, v1 already observed --- in two ``semantic context required'' error cases and its static-analysis-ceiling limitation [Yang 2026, \S{}6.9, L1] --- that some threats turn on intent a rule cannot see; v2 elevates that residue into the central threat-type decomposition and \textit{proves} rules cannot cross it (\S{}5.3). The throughline: v1 was a \textbf{static} interceptor that \textit{named} these gaps; v2 closes them and, in doing so, becomes a \textbf{self-evolving} trust layer --- distilling rules on the lexical side and accruing guarded memory on the semantic side, so it grows cheaper and smarter from its own traffic.

\section*{3. Threat Model and Action Representation}
We consider an AI agent that pursues a user's goal by issuing a stream of \textit{actions} against external systems: a shell, a database, a cloud control plane, a package manager, a CI/CD pipeline, an HTTP client. Each action has real blast radius --- it can delete data, exfiltrate secrets, escalate privilege, or install attacker-controlled code --- so a trust layer must render a per-action verdict \textit{before} the action is executed. This section fixes what an action is, what the trust layer must decide, and the adversary it must decide against. We then introduce the conceptual move that organizes the rest of the paper: a decomposition of agent-action threats by \textit{type}, into \textbf{lexical / fixed-signature} threats that a deterministic rule can resolve and \textbf{semantic / intent-dependent} threats that, we argue, a rule cannot resolve \textit{by construction}.

\subsection*{3.1 Actions and verdicts}
Following v1's formalism [Yang 2026, Def. 1], we model an action as the tuple

\begin{quote}\textbf{a = (type, tool, description, content, params, session, timestamp)},\end{quote}
whose verdict-relevant core is \textbf{(type, tool, content, params)}; the \textbf{session} identifier supports the multi-step (RiskChain) view and the per-session memory of \S{}4, and the \textbf{timestamp} orders a session's stream. We center the discussion on the four core fields. Here \textbf{type} is the channel of effect (e.g., \texttt{shell}, \texttt{db\_\allowbreak{}query}, \texttt{cloud\_\allowbreak{}api}, \texttt{package\_\allowbreak{}install}, \texttt{http\_\allowbreak{}request}, \texttt{file\_\allowbreak{}write}); \textbf{tool} is the specific instrument invoked within that channel (e.g., \texttt{bash}, \texttt{psql}, \texttt{kubectl}, \texttt{aws}, \texttt{npm}, \texttt{curl}); \textbf{content} is the raw, executable payload as the agent emitted it (the literal command line, query string, request body, or file contents); and \textbf{params} is the structured metadata the agent supplies alongside the content (destination host, target resource, scope, flags, credentials referenced). The trust layer sees \texttt{a} exactly as it will be executed --- it is an \textit{interposition} point on the agent's effector path, not a post-hoc log analyzer.

Given \texttt{a} (and the surrounding session context), the trust layer emits one of four \textbf{verdicts}:

\begin{itemize}
\item \textbf{allow} --- execute the action unmodified;
\item \textbf{warn} --- execute, but surface a caution to the operator/log (a soft, fully reversible signal);
\item \textbf{review} --- pause and route to a human (or a stronger automated check) before execution;
\item \textbf{block} --- refuse to execute the action.
\end{itemize}
These verdicts form an \textit{escalation order} \texttt{allow < warn < review < block} in severity. A central design commitment of AgentTrust, motivated below and quantified in the evaluation, is a \textbf{safety invariant on the benign tail}: the system never \textit{hard-blocks} a benign action. When the layer is uncertain about an action it believes is probably legitimate, the worst it may do is escalate to a soft \texttt{warn} or to \texttt{review}; a hard \texttt{block} is reserved for actions it can positively justify as dangerous. The whole system is engineered so that the \textit{cost of a false alarm is paid in friction, never in a broken benign workflow} (\S{}5.5).

\textbf{Adversary.} We assume an adversary who controls action \textit{content} --- either a compromised or jailbroken agent, or, more realistically, an agent steered by indirect prompt injection through the data it ingests (a poisoned web page, issue, file, or tool result). The adversary's aim is to get a harmful action executed. Crucially, the adversary is free to make a harmful action \textit{look} like a routine one: to wrap exfiltration in the same \texttt{curl} an agent uses for telemetry, to read secrets with the same \texttt{kubectl get secret} an engineer uses to debug, to phrase a bulk dump as an ordinary query. The adversary is \textit{not} assumed to have compromised the trust layer itself or the systems it protects; sandbox escape and side channels are out of scope. This threat model --- an adversary who can choose the \textit{surface form} of an action independently of its \textit{intent} --- is precisely what makes a purely lexical defense insufficient, and it is what the taxonomy in \S{}3.2 is built to expose.

\subsection*{3.2 The central decomposition: lexical vs semantic threats}
The key conceptual move of this paper is that agent-action threats split into two structurally different kinds, and that this split --- not the choice of any particular detector --- is what determines which mechanisms can possibly work.

\textbf{Lexical / fixed-signature threats.} Here the danger inheres in a \textit{stable, recognizable token or pattern} in the action's content or params. The malicious action carries a signature that a benign action would essentially never carry: \texttt{rm -rf /\allowbreak{}} over a root path; a reverse shell (\texttt{bash -i >\& /\allowbreak{}dev/\allowbreak{}tcp/\allowbreak{}...}); a hardcoded private key or long-lived credential in source; \texttt{aws iam create-access-key} for a foreign principal; an S3 bucket policy set to \texttt{--acl public-read}; binding a workload to \texttt{cluster-admin}; writing \texttt{/\allowbreak{}etc/\allowbreak{}ld.so.preload}; a request to the cloud metadata IP \texttt{169.254.169.254}; a known-bad download sink. For threats of this kind, a \textit{regex or AST rule can decide the verdict}: the signature is the threat, and matching the signature both detects the danger and (because the benign distribution does not contain that signature) rarely fires on legitimate traffic. Lexical threats are, in a precise sense, \textit{rule-complete}: a sufficiently careful pattern library resolves them with high precision.

\textbf{Semantic / intent-dependent threats.} Here the danger does \textit{not} inhere in any token. The malicious and the benign action \textit{share the same surface form}; the verdict flips only on the \textit{destination, content, scope, or intent} behind that form. The same \texttt{curl} is telemetry to a monitoring endpoint or exfiltration to an attacker; the same \texttt{kubectl get secret} is a debugging read or credential theft; the same database access is a routine \texttt{SELECT} or a bulk \texttt{COPY ... TO} dump shipped off-box; the same dependency line installs a real package or, via a \texttt{postinstall} hook, executes attacker code. For threats of this kind there is, \textit{by construction, no token a rule can key on that separates the malicious instance from its benign twin} --- any pattern broad enough to catch the attack also fires on the legitimate action, and any pattern narrow enough to spare the benign twin also misses the attack. Resolving a semantic threat requires \textit{reasoning about what the action means} --- where it sends data, what it reads, how much, and why --- which is exactly the competence of an LLM judge and exactly what a deterministic matcher lacks. Semantic threats are \textit{rule-blind}: adding rules patches individual fixed-token cases but cannot move the category, because the category's hard cases have no fixed token to patch.

This is the load-bearing claim of the paper, and we make it falsifiable. If semantic threats were merely \textit{under-engineered} rather than \textit{structurally} out of reach, then a determined, hand-authored rule pack should be able to climb the semantic categories. It does not. As our headline result (\S{}5.3, Figure~\ref{fig:fig1}) shows, a cloud/k8s/CI rule pack hand-authored on development data lifts \textit{overall} held-out accuracy from 48\% to 56\%, yet moves the semantic categories by 0pp --- \texttt{data\_\allowbreak{}db} stays at 29, \texttt{observability\_\allowbreak{}benign\_\allowbreak{}hard} stays at 59, and \texttt{supply\_\allowbreak{}chain\_\allowbreak{}pkg} stays at 50 --- while a strong judge carries exactly those same categories (86, 78, and 80 respectively). The improvement the rule pack does deliver is concentrated on the lexical and mixed categories where stable signatures exist (e.g., \texttt{cloud\_\allowbreak{}iam\_\allowbreak{}secrets} 40$\rightarrow$70). The split is therefore not an artifact of effort: it is the boundary between \textit{what is a token} and \textit{what is an intent}.

We deliberately frame the rules-vs-judge argument by threat \textit{type} rather than by any single dataset's accuracy. The structural claim --- rules close the lexical side and cannot cross to the semantic side --- is what the design of AgentTrust stands on; corpus-level accuracy comparisons across distributions are reported in the evaluation and appendix in support of it, not in place of it.

\subsection*{3.3 Twin examples: same surface, opposite verdict}
The intuition pump for the semantic class is the \textit{twin pair}: two actions that a lexical matcher cannot tell apart, with opposite ground-truth verdicts. This twin structure has a v1 pedigree: v1's benchmark was authored so that each scenario is \textit{minimally distinguishable from a benign counterpart} (\texttt{cat README.md} vs \texttt{cat .env}; \texttt{rm -rf /\allowbreak{}node\_\allowbreak{}modules} vs \texttt{rm -rf /\allowbreak{}}) [Yang 2026, \S{}5.1], and v1 already isolated a residue of ``semantic context required'' errors (e.g. \texttt{curl -X PUT -T - \dots{}}) as the LLM-judge's domain [Yang 2026, \S{}6.9]. v2 makes that residue the organizing axis. Table 3.1 gives canonical pairs spanning the threat categories used in our evaluation. Each row is a single \texttt{(type, tool)} surface under which both a benign and a malicious action live; the only difference is in the destination, scope, or intent --- never in a token a rule could safely key on.

\textbf{Table 3.1 --- Twin pairs: identical surface, opposite verdict.}

\begin{table}[ht]\centering\footnotesize
\begin{tabular}{|>{\raggedright\arraybackslash}p{0.23\linewidth}|>{\raggedright\arraybackslash}p{0.23\linewidth}|>{\raggedright\arraybackslash}p{0.23\linewidth}|>{\raggedright\arraybackslash}p{0.23\linewidth}|}\hline
\textbf{(type, tool)} & \textbf{benign twin $\rightarrow$ \textbf{allow}} & \textbf{malicious twin $\rightarrow$ \textbf{block}} & \textbf{what flips the verdict} \\ \hline
\texttt{http\_\allowbreak{}request} / \texttt{curl} & post anonymized metrics to the team's telemetry endpoint & exfiltrate \texttt{\textasciitilde{}/\allowbreak{}.aws/\allowbreak{}credentials} to an attacker host & \textbf{destination} of the egress (and \textit{what} is in the body) \\ \hline
\texttt{http\_\allowbreak{}request} / \texttt{curl} & upload a build artifact to the org's \textit{allow-listed} object store & smuggle a secrets dump through that \textit{same} allow-listed store to a path the attacker controls & \textbf{intent/payload}, not the host --- the host is on the allow-list \\ \hline
\texttt{db\_\allowbreak{}query} / \texttt{psql} & \texttt{SELECT ... LIMIT 100} to answer a user question & \texttt{COPY (SELECT * FROM users) TO PROGRAM 'curl ...'} to dump the table off-box & \textbf{read-vs-extract} semantics and the egress sink inside the query \\ \hline
\texttt{cloud\_\allowbreak{}api} / \texttt{kubectl} & \texttt{kubectl get secret} to debug a failing pod's config & \texttt{kubectl get secret} to harvest credentials for lateral movement & \textbf{purpose/scope} of the read; the command is byte-identical \\ \hline
\texttt{package\_\allowbreak{}install} / \texttt{npm} & add a real, widely-used dependency & add a typosquat whose \texttt{postinstall} hook runs attacker code & \textbf{behavior of the package}, invisible in the install line \\ \hline
\end{tabular}\end{table}
Three of these rows deserve emphasis because they defeat the obvious rule-based defenses:

\begin{itemize}
\item \textbf{Abuse of an allow-listed endpoint} (row 2). A common hardening is an egress \textit{allow-list} of approved destinations. But the malicious twin uses an \textit{approved} host --- the attack rides a sanctioned channel to an attacker-chosen path or payload. Here a destination rule is not merely insufficient; it \textit{licenses} the attack. Only reasoning about \textit{what is being sent and why} separates the pair.
\item \textbf{\texttt{SELECT} vs \texttt{COPY}-exfil} (row 3). One might hope to ban a token like \texttt{COPY ... TO PROGRAM}. But the dangerous behavior --- turning a query into an out-of-band data transfer --- has many surface forms (\texttt{COPY}, \texttt{\textbackslash\{\}copy}, server-side functions, application-level pagination loops), while the benign \texttt{SELECT} shares the same tool and often the same tables. Banning one phrasing neither catches the others nor spares the legitimate query; the verdict turns on \textit{extraction intent and the presence of an egress sink}, not on a keyword.
\item \textbf{\texttt{kubectl get secret} debug vs theft} (row 4). The two actions can be \textit{character-for-character identical}. No transformation of the action's text can distinguish them, because they differ only in \textit{why} the secret is being read. This row is the purest statement of the semantic boundary: when the benign and malicious actions are the same string, no function of that string can separate them.
\end{itemize}
These pairs make concrete why a deterministic floor, however well-tuned, plateaus on the semantic side: its decision is a function of the action's surface, and on a twin pair the surface is constant while the verdict is not.

\subsection*{3.4 Consequences for system design}
The decomposition prescribes the architecture developed in \S{}4 and shown in Figure~\ref{fig:fig4}. Because lexical threats are rule-decidable and cheap, AgentTrust resolves them with a deterministic \textbf{rule floor} and reserves the expensive, reasoning-capable \textbf{judge} for the actions the floor cannot decide --- predominantly the semantic class. Because semantic verdicts depend on intent rather than surface, two further consequences follow. First, a verdict \textit{cache} keyed on the action cannot help on the semantic side: twins that must receive opposite verdicts collapse to the same (or near-identical) key, so cache retrieval cannot separate them --- a negative result we establish directly in \S{}5.4 (Figure~\ref{fig:fig3}) and which is \textit{why} semantic precedent must be stored as guarded retrieval-augmented \textit{reasoning input}, not as a lookup answer. Second, because the system's residual errors live in this hard, intent-dependent tail, the safety invariant of \S{}3.1 --- \textit{never hard-block a benign action} --- is what keeps those errors tolerable: an over-cautious judgment on an ambiguous twin degrades to a \texttt{warn} or \texttt{review}, not to a broken workflow. The self-improving loop of \S{}5.4 (Figure~\ref{fig:fig2}) is built on exactly this split --- the judge teaches the rule floor on the lexical side (cheaper over time) and a guarded memory on the semantic side (smarter over time) --- and we therefore return to the lexical/semantic taxonomy as the organizing axis throughout the evaluation.

\section*{4. System: AgentTrust v2}
AgentTrust v2 is a trust layer that sits between an agent and the world: every consequential action the agent attempts --- a shell command, a cloud API call, a package install, a database query --- is routed through the layer, which returns a verdict in \texttt{\{allow, warn, block, review\}} before the action executes. The design follows directly from the threat taxonomy of \S{}3. Threats split cleanly into two kinds, and the two kinds demand structurally different machinery:

\begin{itemize}
\item \textbf{Lexical / fixed-signature threats} carry their danger in a stable token --- \texttt{rm -rf /\allowbreak{}}, a reverse shell, a hard-coded key, \texttt{iam create-access-key}, \texttt{--acl public-read}, a metadata-service IP. These are \textit{decidable by a pattern matcher}: cheap, deterministic, and high-precision.
\item \textbf{Semantic / intent-dependent threats} share a surface with a benign twin and differ only in destination, content, or scope --- a telemetry \texttt{curl} versus an exfiltration \texttt{curl} (even to an allow-listed endpoint), a debug \texttt{kubectl get secret} versus secret theft, a \texttt{SELECT} versus a \texttt{COPY}-to-exfil. No pattern matcher can separate these without false-positiving on the benign twin; resolving them requires \textit{reading intent}.
\end{itemize}
A single mechanism cannot serve both kinds well. A pattern matcher is the right tool for the lexical class and is structurally blind to the semantic class (quantified in \S{}5.3); a language-model judge reads intent but is more expensive and is wasted on the lexical class, where a regex already gives a correct answer at a fraction of the cost. AgentTrust v2 therefore adopts a \textbf{dual-store architecture} --- a deterministic rule floor for the lexical class and a confidence-gated semantic judge backed by a guarded retrieval memory for the semantic class --- composed by a \textbf{distribution-aware fusion policy} and bounded by a single hard \textbf{safety invariant}. Crucially, the two stores are not static: the judge \textit{teaches} both of them, so the system gets cheaper on the lexical class and smarter on the semantic class as it runs (\S{}4.5). Figure~\ref{fig:fig4} shows the full pipeline and the learning loop.

\textbf{What v2 inherits, and what it adds.} The deterministic core is v1's [Yang 2026]: a \textit{ShellNormalizer} (nine pure-text deobfuscation strategies --- v1's single highest-value component, which alone halves the obfuscated-attack miss rate in ablation), a 42-pattern \textit{ActionAnalyzer}, a 170-rule \textit{PolicyEngine} that already covers cloud IAM (AWS/GCP/Azure), containers, Kubernetes, database privilege, credential files, and anti-forensics, a \textit{SafeFix} remediation engine, a \textit{RiskChain} multi-step detector, a cache-aware \textit{LLM-Judge}, and a \textit{TrustReporter}. v2 keeps this entire core as its \textbf{rule floor} and makes two additions that turn a \textit{static interceptor} into a \textit{self-evolving system}: (i) it promotes the confidence-gated judge from an opt-in fallback to the evaluated core (\S{}4.3), and (ii) it couples floor and judge through a self-improving \textbf{dual-store} --- judge-to-rule distillation on the lexical side, a guarded retrieval memory on the semantic side --- so the system learns from its own decisions (\S{}4.5).

\subsection*{4.1 Action representation and verdicts}
An \textbf{action} is the structured form of an agent tool-call: its \textit{type} (shell, HTTP, cloud-SDK, package, SQL, \dots{}), the \textit{tool} invoked, the \textit{raw content} of the call, and its \textit{parameters}. The layer maps each action to a verdict:

\begin{itemize}
\item \textbf{allow} --- execute the action;
\item \textbf{warn} --- execute, but surface a caution to the operator;
\item \textbf{block} --- refuse the action;
\item \textbf{review} --- suspend the action for human adjudication.
\end{itemize}
\texttt{warn} and \texttt{review} are \textit{soft} outcomes: the benign twin of a semantic threat is never silently destroyed, only escalated for a human's attention. This distinction is what the safety invariant (\S{}4.6) is defined on, and it is what lets the system escalate aggressively on ambiguous traffic without ever catastrophically over-blocking.

\subsection*{4.2 The deterministic rule floor (lexical store)}
The first stage is a \textbf{rule floor}: a set of exact-match and regular-expression rules over the action's content and parameters, each mapping a matched signature to a verdict. The floor is deterministic and high-precision by construction --- a rule fires only on a token it was written to recognize --- which makes it both cheap (no model call) and auditable (every block traces to a named rule). The floor is fronted by v1's \textit{ShellNormalizer}, which rewrites obfuscated payloads to a canonical form (variable expansion, hex/octal escapes, alias resolution, command substitution, adjacent-quote concatenation) so a fixed signature still matches after evasion --- the mechanism that keeps even \textit{obfuscated} lexical threats rule-decidable, and v1's single highest-value component in ablation [Yang 2026, \S{}6.7]. It is the right and sufficient mechanism for the lexical class: a fixed signature such as a reverse-shell payload or a public-read ACL is exactly what a regex decides well.

The floor's \textit{limit} is equally structural and is the pivot of the whole design: because a rule keys on surface tokens, it cannot distinguish two actions that share a surface but differ in intent. Adding rules does not cross this boundary --- a point we establish empirically with a determined steelman in \S{}5.3, where a hand-authored cloud rule pack moves the lexical categories but moves the semantic categories by 0pp. The floor is therefore deliberately \textit{not} asked to resolve the semantic class; that is the judge's job. What the floor \textit{does} gain over time is coverage of \textit{newly observed lexical} threats, via distillation from the judge (\S{}4.5).

\subsection*{4.3 The confidence-gated semantic judge}
Actions the rule floor does not decide are escalated to a \textbf{semantic judge}: a strong language model prompted to read the action in context and return a verdict together with a rationale and a \textbf{confidence}. The judge is the component that reads intent, and it is the component that handles the semantic threats rules cannot --- on the mainly-semantic-attack corpus of \S{}5 it nearly doubles rule accuracy with near-zero false-blocks, and it does so across model families and two providers (\S{}5.2). Recall is not the scarce resource here: even the smallest judge we test catches the semantic attacks the rules miss (\S{}5.5). What scales with model capability is \textit{precision} --- the rate at which the judge over-escalates a benign action.

Because the judge can over-escalate, AgentTrust v2 never acts on a raw judge verdict. The judge's output is admitted only through a \textbf{confidence gate}: an escalation (an allow $\rightarrow$ warn/block/review move) is taken only when the judge's confidence clears a threshold $\tau$; below $\tau$ the action keeps the rule floor's verdict. The gate is the load-bearing control that converts a high-recall but imperfectly-precise judge into a \textit{net-positive} component --- it preserves the judge's catches on the semantic class while suppressing low-confidence over-escalation of benign traffic. Two properties of the gate matter for the system design:

\begin{itemize}
\item \textbf{The threshold is provider-dependent.} Calibration differs across model families: Claude judges are well-calibrated and admit a loose gate, while a GPT-family judge is trigger-happy on ambiguous traffic and requires a tighter gate, but in both cases confidence cleanly separates real catches from over-flags, so a single per-provider $\tau$ recovers a net-positive operating point (\S{}5.2). $\tau$ is a configuration parameter of the deployment, set per judge.
\item \textbf{The gate is what bounds precision cost.} Quantitatively, with a calibrated gate the escalate-only judge is net-positive in-distribution across every model tier we test; the best in-distribution operating point on the independent 630 corpus is a Sonnet-4.6 gated hybrid at \textbf{98.3 / 2.1 / 0.9} (accuracy / FPR / FNR), versus the rule floor's 95.4 / 2.1 / 6.1 --- the judge closes the recall gap with the false-positive rate held flat (\S{}5.2).
\end{itemize}
\subsection*{4.4 The guarded RAG memory (semantic store)}
The semantic class cannot be served by simply \textit{caching} the judge's verdicts. The natural idea --- remember the verdict for an action and reuse it when the action recurs --- fails by construction on exactly this class, because the lookup key is a surface that semantic twins \textit{share}: a verdict-cache keyed on the action retrieves the wrong precedent for the benign-vs-malicious pair it is meant to separate, and has no high-precision operating regime at any coverage (the negative result of \S{}5.4, Figure~\ref{fig:fig3}). The cache fails for the same reason the rule floor fails --- it keys on surface. This memory is the \textit{semantic} generalization of v1's judge cache: v1 shipped an \textbf{exact} content-hash plus incremental-delta cache that amortized token \textit{cost} on identical, recurring contexts [Yang 2026, \S{}4.8]; the semantic store retrieves by \textbf{meaning} for \textit{accuracy}, not by hash for cost --- the same component, evolved from cheaper-on-repeat to smarter-on-semantic.

AgentTrust v2 instead uses retrieval as \textit{evidence for the judge}, not as a replacement for it: a \textbf{guarded RAG memory}. Past adjudications are stored as precedents; when a new semantic action arrives, similar precedents are retrieved and supplied to the judge as additional context, and the judge --- still reading the specific intent of the new action --- produces the verdict. Retrieval here sharpens the judge rather than short-circuiting it, so surface-twins do not collapse to a single answer.

The risk in any self-populated memory is \textit{poisoning}: if the judge's own fallible verdicts are admitted as precedents wholesale, errors compound. We therefore admit precedents through a \textbf{corroboration gate} --- a precedent enters memory only when two independent judges agree on its verdict. This guard is what makes the memory a net gain rather than a liability: with unguarded judge-verdict precedents the retrieval lift is essentially useless and even poisons a category, whereas the corroboration-guarded memory raises semantic accuracy from 70 to \textbf{84} (a +13pp lift) with \textbf{zero} dangerous leaks, and lifts the memory's own correctness from \textbf{70\% to 97\%} (\S{}5.4). The guard admits a \textbf{62\%} corroboration rate --- strict enough to keep memory clean, loose enough to keep it growing.

\subsection*{4.5 The self-improving loop: judge as teacher}
The two stores are coupled by a learning loop in which the \textbf{judge is the teacher} of two students, one per threat type (Figure~\ref{fig:fig4}, learning edges; the online dynamics are Figure~\ref{fig:fig2}):

\begin{itemize}
\item \textbf{Lexical $\rightarrow$ rule distillation (cheaper over time).} When the judge resolves an action whose danger is in fact a \textit{fixed signature} the floor did not yet cover, that signature is distilled into a new deterministic rule. Subsequent occurrences are decided by the floor without a model call, so the judge-call rate falls as the floor's lexical coverage grows. Distillation is applied only to lexically-novel cases; the semantic class is, by construction, \textit{not} ruleifiable and stays with the judge permanently.
\item \textbf{Semantic $\rightarrow$ guarded memory (smarter over time).} Corroborated semantic adjudications are written to the guarded RAG memory (\S{}4.4), so the judge's accuracy on the semantic class improves as the memory warms.
\end{itemize}
An end-to-end online replay over a traffic stream exhibits both effects together: the judge-call rate falls from \textbf{50\% to 44\%} (cheaper, from lexical distillation) while judge-domain accuracy on the non-ruleifiable semantic class rises from \textbf{71\% to 80\%} (smarter, as the memory warms toward its guarded ceiling), with overall accuracy flat because the already-correct ruleifiable cases dominate the mix (\S{}5.4). The loop is a \textit{replay simulation} over measured verdicts, not live production re-judging; we are explicit about this in \S{}6.

\subsection*{4.6 Distribution-aware fusion}
How the rule floor and the judge are \textit{combined} depends on whether the incoming traffic resembles the distribution the rules were authored for. The two stores are fused by a \textbf{distribution-aware} policy:

\begin{itemize}
\item \textbf{In-distribution: rules floor + escalate-only judge.} When the traffic matches the rules' design distribution, the floor is trustworthy on the lexical class, so the floor decides and the judge is allowed only to \textit{escalate} an allow $\rightarrow$ warn/block/review through the confidence gate. This is the net-positive hybrid of \S{}4.3.
\item \textbf{Out-of-distribution: the judge leads.} When the traffic is off the rules' distribution, the floor is no longer a safe foundation --- on the realistic held-out corpus the rule floor not only collapses in accuracy but its false-positive rate climbs to \textbf{22.4\%} (\S{}5.2), so building on it as a floor would \textit{drag down} a capable judge. Here the policy hands leadership to the judge (gated-replace rather than escalate-only): the judge, which lands \textbf{83.6--85.2\%} across the two providers with near-zero false-block on that same corpus, decides and the floor is demoted to an advisory signal.
\end{itemize}
The same two components, recomposed by distribution, give the best of both: the floor's cheap precision where it is valid, and the judge's generalization where it is not.

\subsection*{4.7 The safety invariant}
Above all stages sits one inviolable constraint: \textbf{the system never \textit{hard-blocks} a benign action.} Concretely, no fusion path may turn a truth-benign action into a \texttt{block}; the worst outcome the layer may impose on a benign action is a \textit{soft} \texttt{warn} or \texttt{review} for human attention. Hard-blocks are reserved for actions the system is confident are dangerous. This invariant is what makes aggressive escalation safe to deploy: the cost of the judge being uncertain about a benign action is a caution or a human glance, never the silent destruction of legitimate work.

The invariant holds empirically across every judge $\times$ dataset combination we evaluate (\S{}5.5), and end-to-end across the full online replay --- \textbf{0 benign hard-blocks over 45,000 actions} (\S{}5.4). It is the deployability guarantee on which the rest of the system rests: a trust layer an operator can put inline precisely because its worst-case error on benign traffic is bounded to a soft verdict by construction.

\section*{5. Evaluation}
We evaluate AgentTrust v2 against the three pillars of \S{}1. We first show that the confidence-gated judge handles the semantic threats rules cannot: net-positive in-distribution and dominant on a mainly-semantic-attack corpus, across two model providers (\S{}5.2). We then establish the paper's central structural claim --- that deterministic rules resolve \textit{lexical} threats but cannot cross the \textit{semantic} boundary, no matter how determinedly they are hand-authored (\S{}5.3). Finally we evaluate the self-improving dual-store: why a semantic verdict-cache cannot work, why a \textit{guarded} RAG memory does (+13pp), and an end-to-end online loop that gets cheaper and smarter over a traffic stream while never hard-blocking a benign action (\S{}5.4). \S{}5.5 collects the invariants that hold across every model and dataset.

Our evaluation is designed to answer three questions in order. \textbf{(Q1)} Does a per-action trust decision \textit{generalize} beyond the distribution its policy was written on? \textbf{(Q2)} Is the gap between rules and a judge a matter of \textit{coverage} (write more rules) or of \textit{structure} (a class of threats rules cannot reach)? \textbf{(Q3)} Can a system \textit{learn} from its own judgments --- becoming cheaper on the threats rules can handle and smarter on the ones they cannot --- under a hard safety invariant? We take Q2 to be the load-bearing question and treat it as the headline (\S{}5.3).

\subsection*{5.1 Experimental setup}
\textbf{Datasets.} We use three corpora of agent actions, in increasing order of distance from the policy under test.

\begin{enumerate}
\item \textit{Internal 300} --- the tuning corpus, used during development to set thresholds and rules. We report on it for completeness but never draw a headline claim from it.
\item \textit{Independent 630} --- an out-of-sample corpus held back from tuning. This is our in-distribution test set: same threat taxonomy, not seen while authoring rules or calibrating gates.
\item \textit{External held-out 189} --- a realistic, out-of-distribution (OOD) corpus, validated by an independent second annotator who relabeled it blind (Cohen's $\kappa$ = 0.82 with the original labels) and leakage-checked against v1. This is the true generalization test: it stresses whether a trust decision survives a shift in how actions are phrased and composed, rather than merely repeating the training distribution. It is deliberately adversarial in composition (52\% semantic/attack-heavy), which makes it a conservative setting for any amortization claim (\S{}5.4, \S{}6).
\end{enumerate}
The external corpus spans eight threat categories that we use throughout: \texttt{shell\_\allowbreak{}persistence}, \texttt{fs\_\allowbreak{}config}, \texttt{network\_\allowbreak{}egress}, \texttt{cloud\_\allowbreak{}iam\_\allowbreak{}secrets}, \texttt{cicd\_\allowbreak{}iac}, \texttt{supply\_\allowbreak{}chain\_\allowbreak{}pkg}, \texttt{observability\_\allowbreak{}benign\_\allowbreak{}hard}, and \texttt{data\_\allowbreak{}db}. Each category mixes benign actions with their malicious surface-twins (e.g. a telemetry \texttt{curl} against an allow-listed endpoint vs. an exfiltration \texttt{curl} to the same endpoint; a debugging \texttt{kubectl get secret} vs. credential theft), which is what makes the lexical/semantic distinction operational rather than cosmetic.

\textbf{Provenance and relationship to v1's numbers.} The internal 300 and independent 630 are v1's released benchmarks [Yang 2026, \S{}5]: the 300 spans six categories (50 each; 133 allow / 38 warn / 129 block; \texttt{review} is reserved for ambiguity and is never a ground-truth label), and the 630 is a five-batch adversarial suite (a 30-scenario obfuscation probe, two 100-scenario real-world batches, two 200-scenario expansions; 188 / 98 / 344). Per v1's anti-overfitting protocol the 630 is scored under the \textit{current, patched} ruleset --- an independently-constructed \textit{held-out} set, not a zero-shot one. We re-run the deterministic floor under a stricter \textbf{honest} protocol (per-scenario session reset, and \textbf{excluding} the four synthetic-domain benchmark-compatibility rules v1 isolates in \texttt{benchmark\_\allowbreak{}compat.yaml}), obtaining rules-only \textbf{94.7\%} (300) and \textbf{95.4\%} (630); these reconcile with v1's published \textbf{95.0\% / 96.7\%} (and v1's compatibility-on \textbf{97.0\%}), the small deltas being exactly those compat rules and the session reset. The \textbf{external held-out 189} is \textit{new to this paper}: a realistic corpus built independently of the policy ruleset, leakage-checked against v1, and independently relabeled, blind, by a second annotator who is not an author of this work, reaching Cohen's $\kappa$ = 0.82 with the original labels (meeting v1's recommended $\kappa$ > 0.7 target). It is the true out-of-distribution test.

\textbf{Judges.} We evaluate five judge configurations spanning two providers and a capability ladder: Haiku-4.5, Sonnet-4.6, and Opus-4.8 (Anthropic), GPT-5.5 (OpenAI), and DeepSeek as a deliberately weak baseline (it is the model family v1 shipped with). Using the same prompt and protocol across all five lets us separate provider/calibration effects from raw capability.

\textbf{Rule conditions.} To make the rules-vs-judge comparison fair to the rules, we test three rule configurations: (i) \textit{honest DEFAULT} --- v1's shipped \textbf{170-rule} floor, which \textbf{already} covers cloud IAM (AWS/GCP/Azure), Kubernetes, containers, databases, and credential files [Yang 2026, \S{}4.4]; (ii) \textit{cheat} --- v1's four isolated \textit{benchmark-compatibility} rules (\texttt{benchmark\_\allowbreak{}compat.yaml}, matching synthetic domains such as \texttt{evil.com}, deliberately excluded from production), included to expose the ceiling of in-distribution token tuning; and (iii) \textit{DEFAULT + cloud-pack} --- a determined steelman in which we hand-authored an additional cloud/Kubernetes/CI rule pack on the external \textit{dev} sets and evaluated it held-out. The cloud-pack is the fair test of "just write more rules," and it is central to \S{}5.3.

\textbf{Metrics.} We score all four verdicts honestly (allow / warn / block / review) and report: \textit{exact-verdict accuracy} (acc); \textit{false-positive rate} (FPR) --- a benign action (truth = allow) receiving any non-allow verdict; \textit{false-negative rate} (FNR) --- a dangerous action (truth = block) receiving allow or warn; and the \textit{benign HARD-block} count --- a benign action receiving a hard \texttt{block}. The last metric encodes our safety invariant: a deployable trust layer may at worst inconvenience a benign action with a soft \texttt{warn}/\texttt{review}, but must never hard-block it. We also report the \textit{judge-call rate} and its amortization for the online loop (\S{}5.4). The system architecture under test --- the rule floor, the confidence-gated judge, and the guarded RAG memory arranged as a dual-store --- is summarized in Figure~\ref{fig:fig4} (\S{}4).

\subsection*{5.2 Pillar 1 --- the judge is the component that generalizes}
\textbf{Claim.} A confidence-gated judge is net-positive in-distribution, and out-of-distribution the judge \textit{alone} dominates rules --- and both results hold across model families and providers, not just for a single strong model.

\textbf{Judge alone (Table 1).} In-distribution, deterministic rules are strong on exact-match: the honest rule floor scores 94.7\% (internal 300) and 95.4\% (independent 630), with the residual error concentrated in the subjective \texttt{warn} class. No single judge beats the rule floor on exact-match in-distribution, and we do not claim one does (\S{}7). The picture inverts completely out-of-distribution. On the external 189 corpus the honest rule floor \textit{collapses} to 48.1\% accuracy, with a 22.4\% FPR and a 52.9\% FNR --- i.e. on a realistic distribution shift the rules both over-block benign actions and miss more than half of the dangerous ones. Every \textit{strong} judge, across both providers, lands in a tight 83.6--85.2\% band on exactly that corpus: Opus-4.8 at 85.2 / 1.7 / 4.6, Sonnet-4.6 at 84.1 / 0.0 / 2.3, and GPT-5.5 at 83.6 / 6.9 / 1.1, each with near-zero false-blocks. The judge nearly doubles rule accuracy OOD while \textit{lowering} the false-block rate. DeepSeek, the weak baseline, reaches 69.8\% --- better than rules OOD but with a 31.0\% FPR --- confirming that the generalization is real but that a capable judge is what delivers it cleanly.

The capability ladder is informative as \textit{design justification}, not as the headline. As the judge strengthens (Haiku $\rightarrow$ Sonnet $\rightarrow$ Opus / GPT-5.5), \textit{precision} improves while \textit{recall saturates}: even Haiku-4.5 catches the semantic attacks that rules miss (it reaches 75.7\% OOD with a 0.0\% FPR), so capability buys precision, not recall. This is why the deployable design is a strong, confidence-gated judge rather than a weak one --- but the contribution is the judge's generalization, not "stronger is better."

\textbf{Gated hybrid (Table 2).} In-distribution, the right way to use the judge is as a confidence-gated, escalate-only supplement on top of the rule floor: the rules decide what they decide, and the judge is only allowed to \textit{escalate} an undecided/permissive verdict when it is confident. This is net-positive on every axis. On the independent 630, the rule floor's 95.4 / 2.1 / 6.1 becomes 98.3 / 2.1 / 0.9 with a gated Sonnet-4.6 judge (and 97.9 / 2.1 / 1.5 with Opus-4.8) --- accuracy up, FPR held flat, FNR cut by more than 6$\times$; \textit{all} judge tiers are net-positive here, including Haiku (98.1) and GPT-5.5 (97.5). On the internal 300 the same pattern holds (Opus-4.8 96.7 / 2.3 / 1.6). Out-of-distribution, however, the fusion \textit{inverts}: the rule floor is actively harmful (its 22.4\% FPR drags the hybrid down), so an escalate-only design that trusts the floor underperforms ($\approx$78\%) the judge used \textit{alone} ($\approx$85\%). The correct OOD policy is therefore \textit{judge-leads} (gated-replace) rather than escalate-only. The practical lesson --- and the design we adopt --- is that fusion must be \textit{distribution-aware}: trust the cheap rule floor where it is reliable (in-distribution), and let the judge lead where it is not (OOD). This resolves a tension v1 left open: v1's \textit{naive} second-opinion hybrid (judge every non-block action, take the more severe verdict) cut misses but raised FPR from 2.3\% to 9.0\% and dropped aggregate accuracy to 88.0\% [Yang 2026, Table 9] --- the ``judge hurts'' result --- and v1 explicitly deferred the \textit{asymmetric} fix (escalate only when the rule verdict is uncertain) to future work [Yang 2026, \S{}6.5]. The confidence-gated, escalate-only judge above is precisely that fix, now realized and net-positive across both providers.

\textbf{Why this answers Q1.} A guardrail that only repeats its training distribution is not a trust layer. The judge is the only component here that holds up under a realistic shift, and it does so across two providers --- so the generalization is a property of judge-based decisioning, not an artifact of one vendor's model. (For readers who want it, the rules' in-distribution-to-OOD accuracy \textit{gap} is tabulated in the appendix; we do not headline it, because the structural reason for the gap --- threat type --- is the stronger and more rebuttal-proof claim, and we turn to it next.)

\subsection*{5.3 Pillar 2 --- rules cannot cross the semantic boundary [HEADLINE]}
\textbf{Claim.} Deterministic rules resolve \textit{lexical} (fixed-signature) threats, where danger is a stable token. They cannot resolve \textit{semantic} (intent-dependent) threats, where a benign and a malicious action share the same surface and the verdict depends on destination, scope, or content. This is a \textit{structural} limit, not a coverage gap --- and we prove it with a determined steelman rather than asserting it.

\textbf{The steelman.} The obvious rebuttal to "rules can't catch semantic threats" is "you just didn't write enough rules." So we wrote them --- on top of a floor that \textbf{already} ships 170 production rules covering cloud IAM (AWS/GCP/Azure), Kubernetes, containers, databases, and credential files [Yang 2026, \S{}4.4], so this is not a blank-slate test. We hand-authored a \textit{further} cloud/Kubernetes/CI rule pack directly on the external \textit{dev} sets --- i.e. with full knowledge of the threat families --- and then evaluated it held-out. Figure~\ref{fig:fig1} and Table 3 report the result per category.

The cloud-pack does exactly what rules are good at and nothing more. It lifts overall accuracy from 48\% to 56\%, and the gains land \textit{entirely} on categories with stable signatures: \texttt{shell\_\allowbreak{}persistence} 73 $\rightarrow$ 77, \texttt{fs\_\allowbreak{}config} 50 $\rightarrow$ 58, \texttt{cloud\_\allowbreak{}iam\_\allowbreak{}secrets} 40 $\rightarrow$ 70, \texttt{network\_\allowbreak{}egress} 52 $\rightarrow$ 65, \texttt{cicd\_\allowbreak{}iac} 36 $\rightarrow$ 48. On the categories whose verdict is intent-dependent, the hand-authored pack moves the needle by 0pp: \texttt{data\_\allowbreak{}db} stays at 29, \texttt{observability\_\allowbreak{}benign\_\allowbreak{}hard} stays at 59, and \texttt{supply\_\allowbreak{}chain\_\allowbreak{}pkg} stays at 50. A determined human author, writing rules with the dev distribution in front of them, \textit{cannot} improve the semantic categories at all, because the malicious and benign instances are lexically indistinguishable --- any regex that fires on the attack also fires on its benign twin, so adding the rule trades a false negative for a false positive rather than resolving the case. The \textit{cheat} condition makes the same point from the other side: a maximally token-stuffed rule pack tuned on the test signatures still reaches only 47.6\% overall, because in-distribution rule tuning does not transfer.

\textbf{The judge carries exactly the categories rules cannot.} On those same semantic categories the Opus-4.8 judge reaches 86 (\texttt{data\_\allowbreak{}db}), 78 (\texttt{observability\_\allowbreak{}benign\_\allowbreak{}hard}), and 80 (\texttt{supply\_\allowbreak{}chain\_\allowbreak{}pkg}) --- and 96--100 on the mixed \texttt{cicd\_\allowbreak{}iac}/\texttt{cloud\_\allowbreak{}iam\_\allowbreak{}secrets} categories --- for 85\% overall (Table 3). Figure~\ref{fig:fig1} shows both providers' judges tracking each other across the semantic categories, so the carry is not a single-model fluke. The mechanism is precisely the surface-twin structure: the judge reasons over destination/scope/content, which is the only information that separates the twins, while a rule sees only the shared surface.

\textbf{Why this is the headline.} This decomposition is the structural backbone of the paper. It is rebuttal-proof in a way an accuracy number is not: one cannot answer "rules are blind to semantic threats by construction" with "write more rules," because we \textit{did} write more rules, with the test families in hand, and the semantic categories did not move. The split also tells the system designer exactly where each component belongs --- rules on the lexical floor, the judge on the semantic core --- which is what the dual-store in \S{}5.4 operationalizes.

\subsection*{5.4 Pillar 3 --- a self-improving dual-store [NOVELTY]}
Given the \S{}5.3 split, the system should \textit{learn} asymmetrically: distill the judge's lexical verdicts into the cheap deterministic floor (cheaper over time on the rule-able threats), and accumulate the judge's semantic verdicts into a retrieval memory (smarter over time on the threats rules can never reach). The non-obvious result is that the semantic side cannot be a \textit{cache} --- it must be a \textit{guarded} RAG memory --- and we show why before showing that it works.

\textbf{Lexical $\rightarrow$ rules (cheaper).} On the lexical/fixed-signature categories, the judge acts as a teacher: a confident judge verdict on a previously-undecided lexical action can be distilled into a deterministic rule, so subsequent identical-signature actions are resolved by the cheap floor without a judge call. This is what drives the falling judge-call rate in the online loop below.

\begin{figure}[ht]\centering
\includegraphics[width=\linewidth]{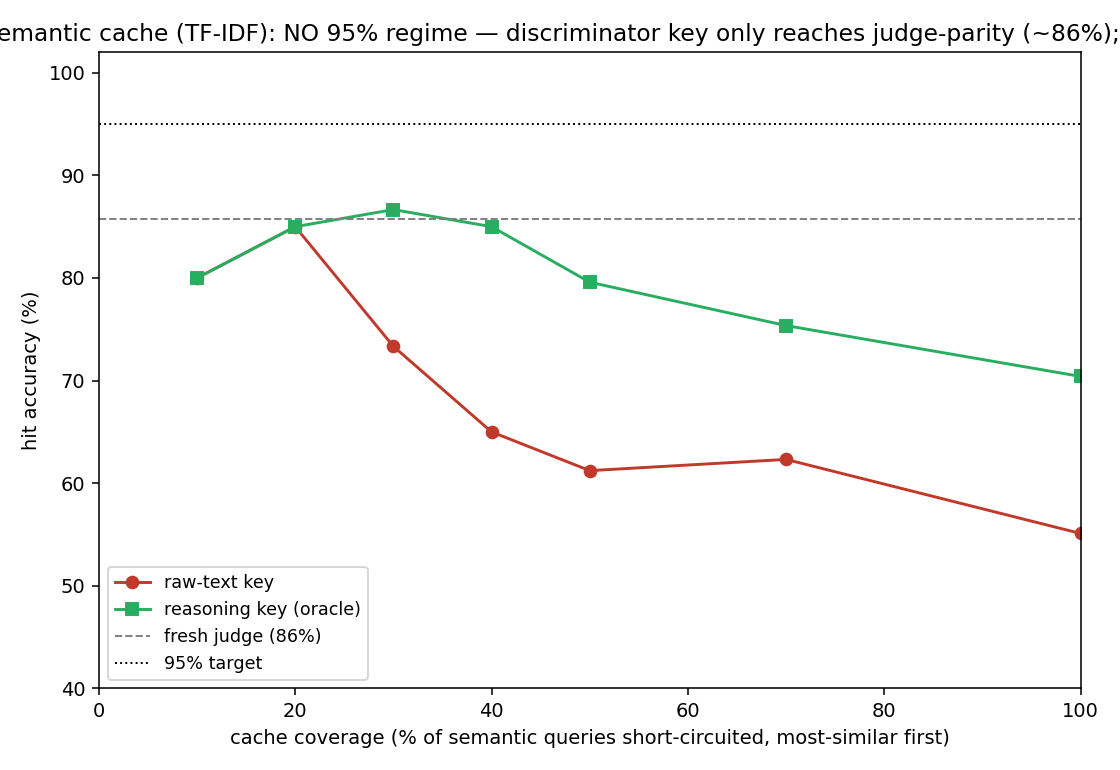}
\caption{The TF-IDF verdict-cache variant: a conservative lower bound that reaches the same conclusion as the neural-retrieval variant -- no high-precision regime exists for a semantic verdict-cache, at any coverage.}\label{fig:figA}
\end{figure}

\textbf{Semantic $\rightarrow$ memory, and why a cache fails (Figure~\ref{fig:fig3}).} The tempting design for the semantic side is a verdict-\textit{cache}: remember the judge's verdict for an action and reuse it when a "similar" action recurs. This cannot work, for exactly the reason rules cannot (\S{}5.3). Table 4 reports the result: retrieval-verdict accuracy on the external semantic slice (n = 98) is $\approx$58\% with raw-text retrieval --- essentially the majority/chance rate --- and a neural (MiniLM) embedding does \textit{not} rescue it. Crucially, there is \textbf{no $\geq$95\% high-precision regime at any coverage}: Figure~\ref{fig:fig3} shows that you cannot trade coverage for precision to carve out a safe high-confidence slice, because retrieval \textit{collapses the surface-twins} --- the benign and malicious instances are nearest neighbors, so the cache returns the wrong precedent precisely on the cases that matter. (The TF-IDF variant in Figure~\ref{fig:figA} is a conservative lower bound and reaches the same conclusion.) This negative result is \textit{why} the semantic store must be retrieval-\textit{augmented} generation, where a judge re-reasons over retrieved precedents, rather than a cache that returns a stored verdict.

\textbf{Semantic $\rightarrow$ guarded RAG (smarter; +13pp).} Using retrieved precedents to \textit{augment} the judge, rather than to replace it, works --- but only with a guard (Table 4). Starting from a no-RAG baseline of 70\% on the semantic slice, RAG with a \textit{clean} memory of ground-truth precedents reaches 79\% (+8pp), establishing a ceiling. But the realistic memory is built from the judge's \textit{own} verdicts, not ground truth; feeding those in \textit{ungated} yields only 72\% (+2pp, effectively useless) and actively \textit{poisons} \texttt{observability\_\allowbreak{}benign\_\allowbreak{}hard} (81 $\rightarrow$ 78) by retrieving the judge's own past mistakes. The fix is a \textit{corroboration guard}: admit a precedent into memory only when two independent judges agree on it (here, a no-RAG Sonnet-4.6 verdict matching Opus-4.8). The guard raises memory correctness from 70\% (ungated) to \textbf{97\%}, at a corroboration rate of 62\%, and lifts the semantic slice to 84\% (+13pp) with \textbf{zero dangerous leaks}. The guarded per-category gains land squarely on the semantic threats: \texttt{cloud\_\allowbreak{}iam} 55 $\rightarrow$ 75, \texttt{data\_\allowbreak{}db} 71 $\rightarrow$ 86, \texttt{observability} 81 $\rightarrow$ 93, \texttt{network} 70 $\rightarrow$ 78. In other words, the memory becomes a poison-resistant, \textit{smarter} semantic store precisely because the guard filters out the judge's unreliable self-labels before they can be retrieved as precedent.

\textbf{End-to-end online loop (Figure~\ref{fig:fig2}).} We tie both stores together in an end-to-end online simulation (Table 5): a Monte-Carlo replay over 30 traffic streams of T = 1500 actions each (45,000 actions total), in which lexical verdicts distill into the rule floor and corroborated semantic verdicts accumulate in the guarded memory as the stream proceeds. Figure~\ref{fig:fig2} shows the two intended dynamics. The system gets \textit{cheaper}: the judge-call rate falls from 50\% (COLD start, empty floor) to 44\% (WARM end) as lexical threats are progressively absorbed into deterministic rules. And it gets \textit{smarter}: accuracy on the judge's domain --- the non-rule-able semantic cases --- rises from 71\% to 80\% as the guarded RAG memory warms toward its 84\% ceiling (\S{}5.4). Overall accuracy is essentially flat (84\% $\rightarrow$ 85\%), which is expected and honest: the rule-able cases that the floor already handles dominate the stream, so the system-level number is insensitive to the semantic improvement even though the semantic \textit{domain} improves by 9pp. Most importantly, the safety invariant holds end to end: \textbf{0 benign HARD-blocks across all 45,000 actions}.

\textbf{Why this answers Q3.} The dual-store learns in the shape that \S{}5.3 demands --- cheaper on the lexical, smarter on the semantic --- and the negative cache result (Figure~\ref{fig:fig3}) is what forces the asymmetric design rather than a single uniform memory. The corroboration guard is what makes self-improvement \textit{safe} rather than self-poisoning, which is the difference between a memory that helps and one that compounds the judge's own errors.

\subsection*{5.5 Invariants and ablations}
Three properties hold across \textit{every} judge and \textit{every} dataset in our study, and together they are what make the system deployable rather than merely accurate.

\textbf{Zero benign hard-blocks.} Across all five judge configurations and all three datasets --- and across the full 45,000-action online run (Table 5) --- no benign action is ever hard-blocked. The worst-case precision cost of the trust layer is a soft \texttt{warn}/\texttt{review}, never a hard \texttt{block}. This is the deployability guarantee: a team can put AgentTrust in the path of real agent traffic knowing the failure mode is friction, not a halted benign workflow.

\textbf{Recall saturates; capability buys precision.} As shown in \S{}5.2, even Haiku-4.5 catches the semantic attacks that rules miss (it reaches 75.7\% OOD at 0.0\% FPR). Moving up the capability ladder improves \textit{precision}/FPR, not recall. This is why a confidence-gated design is the right one --- the gate suppresses the lower-capability models' over-escalation while preserving the recall they already have --- and it is why the gated hybrid is net-positive across all tiers in-distribution (Table 2).

\textbf{The judge generalizes, cross-provider; the gate is load-bearing and provider-dependent.} The OOD dominance of the judge (\S{}5.2, Table 1) holds for both Anthropic and OpenAI models, so it is a property of judge-based decisioning rather than one vendor's calibration. The confidence gate itself is load-bearing: it is what converts a raw judge (which on exact-match in-distribution does \textit{not} beat rules) into a net-positive hybrid. Its threshold, however, is provider-dependent --- different model families separate "confident and correct" from "confident and wrong" at different operating points --- which is a calibration we set by hand and flag as a limitation (\S{}6) rather than a solved problem.

\subsubsection*{Result tables}
All from this session's runs (3 datasets $\times$ 4 model families + DeepSeek + rules). Metric defs: \textbf{acc} = exact-verdict accuracy; \textbf{FPR} = benign(truth=allow) judged non-allow; \textbf{FNR} = dangerous(truth=block) judged allow/warn; \textbf{HARD-block} = benign judged \textit{block} (the safety invariant). Verdicts $\in$ \{allow, warn, block, review\}.

\textbf{Table 1 --- Judge-ALONE (exact-match), acc / FPR / FNR}

\begin{table}[ht]\centering\footnotesize
\begin{tabular}{|>{\raggedright\arraybackslash}p{0.23\linewidth}|>{\raggedright\arraybackslash}p{0.23\linewidth}|>{\raggedright\arraybackslash}p{0.23\linewidth}|>{\raggedright\arraybackslash}p{0.23\linewidth}|}\hline
\textbf{judge} & \textbf{internal 300} & \textbf{independent 630} & \textbf{\textbf{external 189 (OOD)}} \\ \hline
rules-only (honest) & 94.7 / 2.3 / 6.2 & 95.4 / 2.1 / 6.1 & \textbf{48.1 / 22.4 / 52.9} \\ \hline
rules + cloud-pack (steelman) & --- & --- & 56.1 / 22.4 / 33.3 \\ \hline
DeepSeek (weak baseline) & 85.0 / 5.3 / 7.8 & 87.1 / 1.6 / 8.1 & 69.8 / 31.0 / 5.7 \\ \hline
Haiku-4.5 & 84.3 / 6.8 / 4.7 & 80.0 / 0.0 / 12.2 & 75.7 / 0.0 / 11.5 \\ \hline
Sonnet-4.6 & 89.3 / 3.0 / 7.0 & 89.7 / 1.1 / 7.8 & 84.1 / 0.0 / 2.3 \\ \hline
GPT-5.5 & 85.7 / 3.0 / 1.7 & 76.2 / 19.7 / 1.7 & 83.6 / 6.9 / 1.1 \\ \hline
Opus-4.8 & 89.7 / 0.0 / 7.8 & 91.0 / 1.6 / 9.9 & 85.2 / 1.7 / 4.6 \\ \hline
\end{tabular}\end{table}
Reading: in-distribution rules win on exact-match (the gap is the subjective \texttt{warn} class); \textbf{OOD rules COLLAPSE (48\%) while every strong judge across two providers lands 83.6--85.2\% with near-0 false-block.} The judge is the component that generalizes.

\textbf{Table 2 --- Gated HYBRID (rules floor + judge escalate-only), best net-positive point}

\begin{table}[ht]\centering\footnotesize
\begin{tabular}{|>{\raggedright\arraybackslash}p{0.23\linewidth}|>{\raggedright\arraybackslash}p{0.23\linewidth}|>{\raggedright\arraybackslash}p{0.23\linewidth}|>{\raggedright\arraybackslash}p{0.23\linewidth}|}\hline
\textbf{dataset} & \textbf{rules-only (acc / FPR / FNR)} & \textbf{best operating point (acc / FPR / FNR)} & \textbf{fusion policy} \\ \hline
internal 300 & 94.7 / 2.3 / 6.2 & Opus-4.8 96.7 / 2.3 / 1.6 & rules floor + gated escalate-only \\ \hline
independent 630 & 95.4 / 2.1 / 6.1 & Sonnet-4.6 98.3 / 2.1 / 0.9 & rules floor + gated escalate-only \\ \hline
external 189 (OOD) & 48.1 / 22.4 / 52.9 & Opus-4.8 85.2 / 1.7 / 4.6 & judge-led (gated-replace) \\ \hline
\end{tabular}\end{table}
All four judge tiers (Haiku-4.5, Sonnet-4.6, Opus-4.8, GPT-5.5) are net-positive in-distribution; the table lists the single best operating point per dataset (full acc/FPR/FNR). In-distribution the confidence-gated escalate-only judge is net-positive (beats rules on all axes, FPR held). OOD: the rule floor is harmful (22\% FPR) $\rightarrow$ the judge leads (gated-replace, not escalate-only).

\textbf{Table 3 --- Rules steelman, per-category accuracy on external held-out (rules-only $\rightarrow$ +cloud $\rightarrow$ Opus-4.8-judge)}

\begin{table}[ht]\centering\footnotesize
\begin{tabular}{|>{\raggedright\arraybackslash}p{0.23\linewidth}|>{\raggedright\arraybackslash}p{0.23\linewidth}|>{\raggedright\arraybackslash}p{0.23\linewidth}|>{\raggedright\arraybackslash}p{0.23\linewidth}|}\hline
\textbf{category (threat type)} & \textbf{rules-only} & \textbf{+cloud pack} & \textbf{Opus-4.8-judge} \\ \hline
shell\_persistence (lexical) & 73 & 77 & 82 \\ \hline
fs\_config (mixed) & 50 & 58 & 79 \\ \hline
network\_egress (semantic) & 52 & 65 & 83 \\ \hline
cloud\_iam\_secrets (mixed) & 40 & 70 & 100 \\ \hline
cicd\_iac (semantic) & 36 & 48 & 96 \\ \hline
supply\_chain\_pkg (semantic) & 50 & \textbf{50} & 80 \\ \hline
observability\_benign\_hard (semantic) & 59 & \textbf{59} & 78 \\ \hline
data\_db (semantic) & 29 & \textbf{29} & 86 \\ \hline
\textbf{OVERALL} & \textbf{48} & \textbf{56} & \textbf{85} \\ \hline
\end{tabular}\end{table}
Steelman: hand-authored cloud pack (written on dev external sets, tested held-out) lifts overall 48$\rightarrow$56 but \textbf{moves data\_db/observability/supply\_chain by 0pp} --- the semantic boundary is structural. Max-tuned "cheat" rules also = 47.6\% overall (in-dist tuning doesn't transfer).

\textbf{Table 4 --- Semantic memory (external semantic slice, n=98)}

\textit{Verdict-CACHE (dead end):} retrieval-verdict accuracy raw-text $\approx$ 58\% ($\approx$ majority/chance); neural (MiniLM) did not rescue it; \textbf{no $\geq$95\% high-precision regime at any coverage} $\rightarrow$ a semantic verdict-cache keyed on the action is not viable (surface-twins collapse).

\textit{RAG (works, with a guard):}

\begin{table}[ht]\centering\footnotesize
\begin{tabular}{|>{\raggedright\arraybackslash}p{0.23\linewidth}|>{\raggedright\arraybackslash}p{0.23\linewidth}|>{\raggedright\arraybackslash}p{0.23\linewidth}|>{\raggedright\arraybackslash}p{0.23\linewidth}|}\hline
\textbf{condition} & \textbf{semantic acc} & \textbf{$\Delta$ vs no-RAG} & \textbf{dangerous leaks} \\ \hline
no-RAG (baseline) & 70 & --- & 0 \\ \hline
RAG, clean-memory (truth precedents = ceiling) & 79 & +8 & 0 \\ \hline
RAG, realistic (judge-verdict precedents, \textbf{ungated}) & 72 & +2 ($\approx$useless; poisons observability 81$\rightarrow$78) & 0 \\ \hline
RAG, realistic + \textbf{corroboration guard} & \textbf{84} & \textbf{+13} & 0 \\ \hline
\end{tabular}\end{table}
Guard: admit a precedent only where two judges agree (Sonnet-4.6-noRAG == Opus-4.8). Corroboration rate 62\%; memory correctness \textbf{70\% ungated $\rightarrow$ 97\% guarded}. Guarded per-category jumps: cloud\_iam 55$\rightarrow$75, data\_db 71$\rightarrow$86, observability 81$\rightarrow$93, network 70$\rightarrow$78.

\textbf{Table 5 --- Online self-improving loop (Monte-Carlo replay, 30 streams $\times$ T=1500)}

\begin{table}[ht]\centering\footnotesize
\begin{tabular}{|>{\raggedright\arraybackslash}p{0.23\linewidth}|>{\raggedright\arraybackslash}p{0.23\linewidth}|>{\raggedright\arraybackslash}p{0.23\linewidth}|>{\raggedright\arraybackslash}p{0.23\linewidth}|}\hline
\textbf{metric} & \textbf{COLD (start)} & \textbf{WARM (end)} & \textbf{meaning} \\ \hline
judge-call rate & 50\% & \textbf{44\%} & cheaper (lexical$\rightarrow$rule distillation) \\ \hline
judge-domain accuracy (non-rule-able semantic) & 71\% & \textbf{80\%} & smarter (RAG memory warms toward 84) \\ \hline
overall accuracy & 84\% & 85\% & flat (rule-able cases already correct dominate) \\ \hline
benign HARD-blocks & --- & \textbf{0 / 45,000} & safety invariant held end-to-end \\ \hline
\end{tabular}\end{table}
\textit{Universal invariants (across all judges $\times$ datasets):}

\begin{itemize}
\item \textbf{0 benign HARD-blocks} --- worst-case precision cost is a soft \texttt{warn}, never a hard block.
\item \textbf{Recall saturates} --- even Haiku catches the semantic attacks rules miss; capability buys \textit{precision}, not recall.
\item \textbf{Gated hybrid net-positive in-distribution}; \textbf{judge generalizes OOD, cross-provider}.
\end{itemize}
\section*{6. Limitations}
We have argued that trust for agent actions decomposes cleanly by threat \textit{type} --- that \textit{lexical}, fixed-signature threats are rule-decidable while \textit{semantic}, intent-dependent threats are out of reach for deterministic rules by construction (Figure~\ref{fig:fig1}), and that a confidence-gated judge plus a guarded dual-store turns this decomposition into a system that becomes cheaper on the lexical class and smarter on the semantic class over time (Figure~\ref{fig:fig2}). That decomposition is the structural claim we stand on, and it is what survives the limitations below. The results that fill it in, however, were obtained under conditions we state plainly here, because the honesty of this accounting is what makes the decomposition credible rather than convenient.

\textbf{Evaluation is on curated corpora, not real production traffic.} All numbers in this paper come from an agent/replay harness run over three curated corpora --- an internal tuning set of 300 actions, an independent out-of-sample set of 630, and an external held-out set of 189 actions (relabeled blind by an independent second annotator, $\kappa$ = 0.82; leakage-checked against v1). The external set is realistic and adversarially relabeled, and it is on this set that rules collapse to 48.1\% accuracy while the four strong judges across two providers cluster at 83.6--85.2\% with near-zero false-blocks --- the central generalization result. But a curated, leakage-checked corpus is still not a live production stream: real agent traffic has a different category mix, a different base rate of dangerous actions, distribution drift over time, and adversaries who adapt to the deployed guard. We make no claim that the absolute accuracies transfer unchanged to production; what we claim transfers is the \textit{ordering} --- that semantic categories move 0pp under a determined rule steelman (data\_db 29$\rightarrow$29, observability 59$\rightarrow$59, supply\_chain 50$\rightarrow$50, Table 3) while the judge carries exactly those categories (78--100\%). Relatedly, the judge verdicts evaluated here are model-generated for this study; in deployment, judge latency, cost, and refusal behavior on adversarial inputs would all need to be measured against live traffic rather than assumed from the corpus.

\textbf{The online loop is a replay simulation, not a live or production system.} The end-to-end self-improving result (Figure~\ref{fig:fig2}; Table 5) is a Monte-Carlo replay over 30 streams of 1,500 actions each. Its dynamics are driven by \textit{measured} verdicts --- a cold (no-RAG) and a warm (corroboration- guarded) measurement, gated by the real memory-admission and rule-distillation logic --- rather than by re-invoking the judge live on every streamed action. This is the right design for a controlled, reproducible study of the loop's \textit{trajectory} (judge-call rate falling 50\%$\rightarrow$44\%, judge-domain accuracy rising 71\%$\rightarrow$80\%, and 0 benign hard-blocks across 45,000 actions), but it is not a live re-judging system and it is not the production \texttt{src/\allowbreak{}} code path. A deployed loop would have to contend with concurrency, online label noise, and the feedback risk that the distilled rule floor and the RAG memory both learn from the same judge --- effects a replay over fixed verdicts cannot fully exercise.

\textbf{The amortization magnitude is corpus-conservative.} The headline of the loop is \textit{direction}, not size: the drop in judge-call rate (50\%$\rightarrow$44\%) is small precisely because our external corpus is 52\% semantic/attack-heavy, so most actions legitimately require the judge and cannot be retired to the rule floor. On real, benign-heavy traffic --- where the overwhelming majority of actions are either obviously safe or hit a stable lexical signature --- the amortized judge cost would fall much further. We therefore report the 50\%$\rightarrow$44\% figure as a conservative lower bound on the cheapening effect, not as the expected production saving, and we do not extrapolate a dollar or latency figure from it.

\textbf{The strongest semantic-memory result depends on a single, costly guard variant.} Our best RAG result --- +13pp on the semantic slice (70$\rightarrow$84), with memory correctness rising 70\%$\rightarrow$97\% and zero dangerous leaks (Table 4) --- is achieved by a \textit{corroboration} guard that admits a precedent only when two judges agree (here Sonnet-4.6-noRAG == Opus-4.8). This is what separates the guarded result from the realistic ungated one, which is nearly useless (+2pp) and actually poisons the observability category (81$\rightarrow$78). The guard works, but it costs a second judge call at admission time, its corroboration rate is 62\% (so 38\% of candidate precedents are discarded), and it is the \textit{only} guard variant we evaluated. We do not claim it is optimal; cheaper or higher-yield admission policies (single-judge confidence gating, abstention, human spot-checks) are unexplored.

\textbf{The figures are from one guard configuration, modest N, and hand-set gate thresholds.} The external held-out set is 189 actions (98 in the semantic slice used for the memory experiments), which is large enough to separate a 48\% rule floor from an 84--85\% judge but small enough that per-category accuracies carry non-trivial confidence intervals; we report point estimates and do not over-read single-category differences of a few points. The negative cache result (Figure~\ref{fig:fig3}) is similarly measured on this slice --- it shows a verdict-cache stalls at $\approx$58\% with no $\geq$95\% high-precision regime at any coverage, which is \textit{why} the semantic store must be RAG rather than a cache, but it is one corpus, not a proof. Finally, the confidence-gate thresholds that make the hybrid net-positive in-distribution are \textbf{provider-dependent and hand-set}, not auto-calibrated: the gate is load-bearing (it is what holds FPR flat while raising accuracy, e.g. Sonnet-4.6 98.3 / 2.1 / 0.9 on the 630 set), and a deployment on a new model or provider would need to re-tune it. Auto-calibrating the gate is left to future work.

We also state the boundaries of our argument as plainly as its content, to forestall misreading:

\begin{itemize}
\item \textbf{We do not claim the judge alone beats rules in-distribution.} On exact-verdict match, in-distribution rules are competitive or better (rules 95.4 vs. the strongest judge-alone 91.0 on the 630 set, Table 1); the residual gap is largely the subjective \texttt{warn} class. The in-distribution win is the \textit{gated hybrid} (rules floor + escalate-only judge), not the judge in isolation. The judge-alone advantage is an \textit{out-of-distribution} phenomenon.
\item \textbf{We do not claim that "adding rules" is a contribution, nor that more rules close the gap.} The opposite is our point. A determined, hand-authored cloud rule pack --- written on dev external sets and tested held-out --- lifts overall external accuracy only 48$\rightarrow$56\% and moves the semantic categories by 0pp (Table 3); a maximally token-tuned "cheat" rule set still scores 47.6\% out-of-distribution. The boundary is structural, not an artifact of insufficient engineering effort.
\item \textbf{We do not claim a deployed or production system, nor a real-traffic evaluation.} Every result is from the replay harness over curated corpora, as detailed above. The self-improving loop is demonstrated, not shipped.
\item \textbf{We do not headline the rules' internal-vs-external accuracy gap.} A naive reading of that gap invites the rebuttal that the benchmark is simply "too hard" or that one should "write more rules." We deliberately keep that gap in the appendix and stand instead on the \textit{threat-type} claim --- that the semantic categories are unreachable by rules regardless of effort --- which is the structural, rebuttal-resistant version of the same observation.
\item \textbf{We do not claim universality of the absolute numbers across models or providers.} What generalizes, on our evidence, is the \textit{shape}: across Haiku-4.5, Sonnet-4.6, Opus-4.8, and GPT-5.5, recall on the semantic attacks that rules miss saturates even at the weakest tier, while capability buys \textit{precision} (lower FPR), not recall --- and a gated hybrid is net-positive in-distribution while the judge leads out-of-distribution. We claim the structure, not the decimals.
\end{itemize}
\section*{7. Conclusion}
Trust for AI-agent actions splits cleanly by threat type, and that split is the organizing principle of this work. \textit{Lexical}, fixed-signature threats --- \texttt{rm -rf /\allowbreak{}}, a hardcoded key, a known-bad sink --- are decidable by deterministic rules and should be: rules are cheap, exact, and auditable. \textit{Semantic}, intent-dependent threats --- a telemetry \texttt{curl} versus an exfiltration \texttt{curl} to an allow-listed endpoint, a debugging \texttt{kubectl get secret} versus a theft, a benign dependency versus a \texttt{postinstall} RCE --- share their surface with a benign twin and are out of reach for rules by construction; a determined rule steelman moves these categories 0pp (Table 3). A strong LLM judge is the component that crosses this boundary: on a mainly-semantic-attack, leakage-checked corpus it nearly doubles rule accuracy (rules 48.1\% $\rightarrow$ judges 83.6--85.2\%) with near-zero false-blocks, and the effect holds across two model providers (Figure~\ref{fig:fig1}).

We turned this decomposition into a \textit{self-evolving} system. A confidence-gated judge supplies recall on the semantic class without disturbing in-distribution precision (gated hybrid 98.3 / 2.1 / 0.9, Sonnet-4.6 on the 630 set, Table 2); a guarded dual-store then makes the system \textit{learn} from its own judgments --- the judge distills a growing deterministic rule floor on the lexical class (cheaper), and a corroboration-guarded RAG memory accrues semantic precedent (smarter; +13pp, 70$\rightarrow$84, memory correctness 70\%$\rightarrow$97\%, Table 4) --- while a verdict-cache provably cannot, because surface-twins collapse under retrieval (Figure~\ref{fig:fig3}). End to end, an online replay shows the judge-call rate falling (50\%$\rightarrow$44\%), judge-domain accuracy rising (71\%$\rightarrow$80\%), and --- across all judges, datasets, and the full 45,000-action stream --- zero benign hard-blocks (Figure~\ref{fig:fig2}; Table 5). That last invariant is the deployability guarantee: the worst case the system can inflict on a legitimate action is a soft \texttt{warn} or a review request, never a hard block.

The honest scope of these results is a replay study over curated corpora, with hand-set gate thresholds and a single, costly memory guard (\S{}6). The contribution is not a shipped product but a characterization and a design: that the rules-versus-judge question is settled by threat type rather than by effort, that the judge is what makes agent-action trust \textit{generalize} beyond the rules' distribution, and that a guarded, self-improving dual-store can get cheaper on the lexical and smarter on the semantic over time without ever hard-blocking a benign action. The clearest paths forward follow directly from the limitations: a real-traffic evaluation, an auto-calibrated confidence gate, additional memory-admission guards, and a productionized online loop.

\section*{References}
\begin{itemize}
\item Aamodt, A. and Plaza, E. (1994). \textit{Case-Based Reasoning: Foundational Issues, Methodological Variations, and System Approaches.} AI Communications, 7(1), 39--59.
\item Bai, Y., Kadavath, S., Kundu, S., Askell, A., Kernion, J., Jones, A., Chen, A., Goldie, A., et al. (2022). \textit{Constitutional AI: Harmlessness from AI Feedback.} arXiv:2212.08073.
\item Bühler, C., Biagiola, M., Di Grazia, L., and Salvaneschi, G. (2026). \textit{AgentBound: Securing Execution Boundaries of AI Agents.} In Proc. 34th ACM Joint European Software Engineering Conference and Symposium on the Foundations of Software Engineering (FSE 2026), article FSE096. ACM. \url{https://doi.org/10.1145/3808103}
\item Debenedetti, E., Zhang, J., Balunović, M., Beurer-Kellner, L., Fischer, M., and Tramèr, F. (2024). \textit{AgentDojo: A Dynamic Environment to Evaluate Prompt Injection Attacks and Defenses for LLM Agents.} NeurIPS 2024 Datasets and Benchmarks Track. arXiv:2406.13352.
\item Debenedetti, E., Shumailov, I., Fan, T., Hayes, J., Carlini, N., Fabian, D., Kern, C., Shi, C., Terzis, A., and Tramèr, F. (2025). \textit{Defeating Prompt Injections by Design (CaMeL).} arXiv:2503.18813.
\item Dong, Y., Mu, R., Zhang, Y., Sun, S., Zhang, T., Wu, C., Jin, G., Qi, Y., Hu, J., Meng, J., Bensalem, S., and Huang, X. (2024). \textit{Safeguarding Large Language Models: A Survey.} arXiv:2406.02622.
\item Grattafiori, A., et al. (Llama Team, AI @ Meta) (2024). \textit{The Llama 3 Herd of Models} (introduces Llama Guard 3). arXiv:2407.21783.
\item Greshake, K., Abdelnabi, S., Mishra, S., Endres, C., Holz, T., and Fritz, M. (2023). \textit{Not What You've Signed Up For: Compromising Real-World LLM-Integrated Applications with Indirect Prompt Injection.} arXiv:2302.12173. (AISec 2023.)
\item Hinton, G., Vinyals, O., and Dean, J. (2015). \textit{Distilling the Knowledge in a Neural Network.} arXiv:1503.02531.
\item Inan, H., Upasani, K., Chi, J., Rungta, R., Iyer, K., Mao, Y., Tontchev, M., Hu, Q., Fuller, B., Testuggine, D., and Khabsa, M. (2023). \textit{Llama Guard: LLM-based Input-Output Safeguard for Human-AI Conversations.} arXiv:2312.06674.
\item Lewis, P., Perez, E., Piktus, A., Petroni, F., Karpukhin, V., Goyal, N., Küttler, H., Lewis, M., Yih, W., Rocktäschel, T., Riedel, S., and Kiela, D. (2020). \textit{Retrieval-Augmented Generation for Knowledge-Intensive NLP Tasks.} NeurIPS 2020. arXiv:2005.11401.
\item Meta Prompt Guard. \textit{Prompt-Guard-86M / Llama-Prompt-Guard-2} (mDeBERTa-based prompt-injection \& jailbreak classifier, released with Llama 3.1 / Llama 4). Meta / Hugging Face model card \texttt{meta-llama/\allowbreak{}Prompt-Guard-86M}.
\item Open Policy Agent (OPA). \textit{Open Policy Agent and the Rego policy language.} CNCF graduated project. \url{https://www.openpolicyagent.org/}
\item OWASP (2025). \textit{OWASP Top 10 for Large Language Model Applications (2025).} OWASP GenAI Security Project. \url{https://genai.owasp.org/resource/owasp-top-10-for-llm-applications-2025/}
\item ProtectAI. \textit{deberta-v3-base-prompt-injection-v2} (DeBERTa-v3 prompt-injection detector). Hugging Face model \texttt{protectai/\allowbreak{}deberta-v3-base-prompt-injection-v2}.
\item Rebedea, T., Dinu, R., Sreedhar, M. N., Parisien, C., and Cohen, J. (2023). \textit{NeMo Guardrails: A Toolkit for Controllable and Safe LLM Applications with Programmable Rails.} EMNLP 2023 (System Demonstrations), ACL Anthology 2023.emnlp-demo.40. arXiv:2310.10501.
\item Rebuff. \textit{Rebuff: A self-hardening prompt-injection detector} (heuristics + LLM classifier + vector-DB of past attacks). \url{https://github.com/protectai/rebuff}
\item Ruan, Y., Dong, H., Wang, A., Pitis, S., Zhou, Y., Ba, J., Dubois, Y., Maddison, C., and Hashimoto, T. (2024). \textit{Identifying the Risks of LM Agents with an LM-Emulated Sandbox (ToolEmu).} ICLR 2024 (Spotlight). arXiv:2309.15817.
\item Shalev-Shwartz, S. (2012). \textit{Online Learning and Online Convex Optimization.} Foundations and Trends in Machine Learning, 4(2), 107--194.
\item Shi, T., He, J., Wang, Z., Li, H., Wu, L., Guo, W., and Song, D. (2025). \textit{Progent: Programmable Privilege Control for LLM Agents.} arXiv:2504.11703.
\item Vigil. \textit{Vigil: detect prompt injections, jailbreaks, and other risky LLM inputs} (YARA signatures, transformer classifiers, and vector similarity). \url{https://github.com/deadbits/vigil-llm}
\item West, P., Bhagavatula, C., Hessel, J., Hwang, J. D., Jiang, L., Le Bras, R., Lu, X., Welleck, S., and Choi, Y. (2021). \textit{Symbolic Knowledge Distillation: from General Language Models to Commonsense Models.} NAACL 2022. arXiv:2110.07178.
\item Yang, C. (2026). \textit{AgentTrust: Runtime Safety Evaluation and Interception for AI Agent Tool Use.} arXiv:2605.04785. (v1 of this system.)
\item Zheng, L., Chiang, W., Sheng, Y., Zhuang, S., Wu, Z., Zhuang, Y., Lin, Z., Li, Z., Li, D., Xing, E. P., Zhang, H., Gonzalez, J. E., and Stoica, I. (2023). \textit{Judging LLM-as-a-Judge with MT-Bench and Chatbot Arena.} NeurIPS 2023. arXiv:2306.05685.
\end{itemize}
\end{document}